\documentclass[runningheads]{llncs}
\usepackage{amsmath}
\usepackage{amssymb}
\usepackage{booktabs} % For pretty tables
\usepackage{caption} % For caption spacing
\usepackage{subcaption} % For sub-figures
\usepackage{graphicx}
\usepackage{pgfplots}
\usepackage[all]{nowidow}
\usepackage[utf8]{inputenc}
\usepackage{tikz}
\usetikzlibrary{er,positioning,bayesnet}
\usepackage{multicol, multirow}
\usepackage{algpseudocode,algorithm,algorithmicx}
\usepackage[flushleft]{threeparttable}
\usepackage{tabularx}
\usepackage{soul}
\newcolumntype{Y}{>{\centering\arraybackslash}X}

\makeatletter
\newcommand{\@chapapp}{\relax}%
\makeatother

\usepackage[title,toc,titletoc,header]{appendix}

\usepackage{enumerate}
\usepackage[shortlabels]{enumitem}

\usepackage{hyperref}

\usepackage{fullpage}
\usepackage[numbers,sort&compress,super]{natbib}
\usepackage{hyperref}
\usepackage[capitalize]{cleveref}
\usepackage{setspace}

\makeatletter
\renewcommand*\l@author[2]{}
\renewcommand*\l@title[2]{}
\makeatletter

\usepackage{lineno}
\newcommand{\comment}[1]{}

\definecolor{blue}{HTML}{1F77B4}
\definecolor{orange}{HTML}{FF7F0E}
\definecolor{green}{HTML}{2CA02C}

\pgfplotsset{compat=1.14}

\numberwithin{equation}{section} % equation label formatting

\DeclareMathOperator*{\argmin}{arg\,min}
\DeclareMathOperator{\speed}{\texttt{Efficiency}}
\DeclareMathOperator{\sape}{SAPE}
\DeclareMathOperator{\eff}{\texttt{Effectiveness}}
\DeclareMathOperator{\paramcon}{\texttt{Consistency}_{\theta}}
\DeclareMathOperator{\outputcon}{\texttt{Consistency}_{y}}
\DeclareMathOperator{\conkl}{\texttt{Consistency}_{KL}}
\DeclareMathOperator{\kl}{\text{KL}}

\DeclareMathOperator{\certdis}{\texttt{Certifiability}}
\DeclareMathOperator{\pr}{\text{Pr}}
\DeclareMathOperator{\len}{\text{len}}

\newcommand{\test}{\text{test}}

\newcommand{\init}{\text{init}}
\DeclareMathOperator{\poly}{\text{poly}}

\newcommand{\del}{\text{del}}
\newcommand{\rem}{\text{rem}}
\newcommand{\pred}{\text{pred}}
\newcommand{\node}{\text{node}}
\newcommand{\rmax}{\text{rmax}}

\setcitestyle{authoryear,round,citesep={;},aysep={,},yysep={;}}
\algblock{Input}{EndInput}
\algnotext{EndInput}
\algblock{Output}{EndOutput}
\algnotext{EndOutput}

\begin{document}
\sloppy
\title{An Introduction to Machine Unlearning}
%
%\titlerunning{Abbreviated paper title}
% If the paper title is too long for the running head, you can set
% an abbreviated paper title here
%

\author{Salvatore Mercuri\inst{1} \and Raad Khraishi\inst{1}\fnmsep\inst{2}\and Ramin Okhrati\inst{2} \\ Devesh Batra\inst{1} \and Conor Hamill\inst{1} \and Taha Ghasempour\inst{1}\fnmsep\thanks{No longer at institute\inst{1}}\and Andrew Nowlan\inst{1}}
\institute{Data Science \& Innovation, NatWest Group, London, United Kingdom\thanks{Correspondence to: \email{salvatore.mercuri@natwest.com}} 
 \and  Institute of Finance and Technology, UCL, London, United Kingdom\thanks{Correspondence to: \email{raad.khraishi@ucl.ac.uk}}
}

%
%\authorrunning{F. Author et al.}
% First names are abbreviated in the running head.
% If there are more than two authors, 'et al.' is used.
%
%
\maketitle              % typeset the header of the contribution
\begin{abstract}

Removing the influence of a specified subset of training data from a machine learning model may be required to address issues such as privacy, fairness, and data quality.
Retraining the model from scratch on the remaining data after removal of the subset is an effective but often infeasible option, due to its computational expense.
The past few years have therefore seen several novel approaches towards efficient removal, forming the field of ``machine unlearning", however, many aspects of the literature published thus far are disparate and lack consensus.
In this paper, we summarise and compare seven state-of-the-art machine unlearning algorithms, consolidate definitions of core concepts used in
the field, reconcile different approaches for evaluating algorithms, and discuss issues related to applying machine unlearning in practice.

\end{abstract}

\tableofcontents
\comment{\newpage
\noindent\textbf{Notation.}

\begin{itemize}
	\item $\mathcal{X} = \mathbb{R}^p$: feature space; $1\leq p\in\mathbb{Z}$: number of attributes.
	\item $\mathcal{Y}$: label space; for classification tasks $\mathcal{Y} = \{0, 1\}^c$, where $1\leq c\in\mathbb{Z}$ is the number of classes. 
	\item $\mathcal{Z} = \mathcal{X}\times\mathcal{Y}$: data domain; $\mathcal{Z}^*$: space of multisets of elements of $\mathcal{Z}$.
	\item $D\in\mathcal{Z}^*$: initial training data; $D_{\del}$: data to be unlearned; $D_{\rem} = D\setminus D_{\del}$: remaining data.
	\item $n$: number of data points in $D$; $m$: number of data points in $D_{\del}$; $p$: dimensionality of $D$.
	\item $\mathrm{\Theta}$: model space; if the model is parametric $\mathrm{\Theta}\subseteq\mathbb{R}^p$.
	\item $\mathcal{A}:\mathcal{Z}^*\to\mathrm{\Theta}$: a randomised training procedure.
	\item $\mathcal{U}:\mathrm{\Theta}\times\mathcal{Z}^*\times\mathcal{Z}\to\mathrm{\Theta}$: machine unlearning procedure.
	\item $\mathcal{C}$: certificate of unlearning.
	\item $\mathcal{U}^*$: na\"{i}ve unlearning procedure.
	\item $\theta$: original trained trained model, $\mathcal{A}(D)$; $\theta^{\mathcal{U}}$: unlearned model; $\theta^*$: na\"ive unlearned model.
	\item $L:\theta\times\mathcal{Z}^*\to\mathbb{R}$: the loss function; empirical loss on $D$ expressed as $L(\theta, D) = \frac{1}{n}\sum_{i=1}^n L_i(\theta)$.
\end{itemize}}

\newpage

\section{Introduction}
\label{introduction}
% TODO: Add citations

% Why is it important? Remove influence of a data point. Regulation, privacy, data quality.
% What is done today? Naive retraining and machine unlearning
% More motivating examples? E.g., specific regulations.

Widely used machine learning algorithms are able to learn from new data using batch or online training methods but are incapable of efficiently adapting to data removal.
Data removal, however, may be required to address various issues around privacy, fairness, and data quality.
For example, the ``Right to be Forgotten" in the European Union’s General Data Protection Regulation (GDPR) provides individuals with the right to request the removal of their data from an organisation’s records.
Though data may be removed from databases, any machine learning model previously trained on the removed data will retain its information, and proposed legislation such as the European Union Artificial Intelligence Act\footnote{\href{https://artificialintelligenceact.eu/the-act/}{https://artificialintelligenceact.eu/the-act/}} may require further action.
Beyond privacy, other applications include data correction by removing erroneous data \citep{biggio2013poisoning}, bias reduction \citep{quenouille1956}, model explainability \citep{doshivelez2017rigorous}, and uncertainty quantification \citep{shafervovk2008conformal}.
One way to remove the information of deleted data from trained models is to remove the specified data from the training dataset and then retrain the model from scratch on the remaining data.
Though this ``na\"{i}ve" retraining approach is effective in removing the influence of the removed data, it is often inefficient and may entail large computational costs.
For example, modern deep learning models may take several weeks to train and can cost millions of dollars.\footnote{\href{https://lambdalabs.com/blog/demystifying-gpt-3/}{https://lambdalabs.com/blog/demystifying-gpt-3/}}

The field of ``machine unlearning" addresses the problem of removing subsets of training data, through the development of algorithms that guarantee the removal of the subset data's information from the trained model more efficiently than na\"ive retraining.
The term machine unlearning originates from \citet{CauYang}, in which they develop one of the first systematic unlearning algorithms.
The theory has since seen a number of proposed approaches which can broadly be categorised as exact or approximate unlearning.
Exact unlearning algorithms reduce the large computational cost of na\"{i}ve retraining by structuring the initial training so as to allow for more efficient retraining; in doing so they replicate the same model that would have been produced under na\"{i}ve retraining.
In contrast, approximate unlearning algorithms avoid the need for full retraining, speeding up the process of unlearning by allowing a degree of approximation between the output model and the na\"ive retrained model.
Approximate methods typically leverage at least one of the following in order to unlearn: information about the
data to be removed \citep{Golatkar_2019}, information about the remaining data after removal \citep{guo2020certified, he2021deepobliviate}, or information cached during the original machine learning training \citep{Wu_2020, he2021deepobliviate}.

The evaluation approach of an unlearning algorithm often depends on the application and the category of the unlearning approach, resulting in evaluation frameworks and measures that are inconsistent throughout the literature.
In particular, there is little consensus in measuring how well an unlearning algorithm has removed the information of the deleted data.
These inconsistencies even extend to core definitions used in the field including what defines an unlearning algorithm.
The prior work of \citet{Mahadevan_2021} offers a benchmarked comparison between different machine unlearning methods and is a key review paper, however, it only considers three approximate methods with a focus on linear classification methods trained with gradient descent.
Moreover, there is a lack of discussion around applying machine unlearning techniques in practice.
%For example, while in theory we may evaluate unlearning by comparing it to the baseline na\"ive retraining, it is generally unclear how to evaluate and monitor unlearning in practice since computing the na\"ive retrained model defeats the purpose of unlearning.

In the present paper, we first provide standardised definitions of concepts which occur throughout the literature in \Cref{preliminaries}.
The measures used to evaluate unlearning methods are collated in \Cref{sec:evaluation} and we discuss their applicabilities and limitations. \Cref{sec:exact_methods} and \Cref{sec:approx_methods} review a total of seven state-of-the-art unlearning methods --
exact and approximate respectively -- and summarise their scope, benefits and limitations.
Finally, in \Cref{sec:discussion}, we bring together our findings from this review, providing comparisons among the different unlearning techniques.
We also discuss aspects of taking unlearning theory into practice, by giving proposals for algorithm selection and monitoring procedures.

\paragraph{Selection criteria.}

We select seven methods that unlearn data points from machine learning models, covering a broad scope of applicability across both exact and approximate unlearning, so that we include at least one method that may be applied to gradient-descent based models, decision-tree based models, those with convex loss functions and those with non-convex loss functions. We aim to avoid including more than one unlearning algorithm that covers the same or highly similar applicability, unlearning type, and algorithmic methodology.

These selection criteria naturally mean that many works are excluded, but not from lack of merit or significance, and here we mention their achievements. In a relatively early contribution, \citet{Ginart_2019} develop an exact unlearning algorithm applied to the $k$-means algorithm and in the appendix they are the first to propose the notion of approximate unlearning. 
This work is significant for inspiring a strain of methods based around statistical approximation and indistinguishability, of which we include further developments in this paper, most directly by \citet{guo2020certified, Neel_2020}. \citet{Schelter_2021} develop a method applicable to decision-tree based models; this has similar applicability and similarly strong empirical results as \cite{Brophy_2021}, which is included in this paper. 
In \citet{coded_2021}, the authors develop an exact unlearning method that uses a similar sharding methodology as seen in SISA, \Cref{sec:sisa}. 
\citet{Gupta_2021} highlight issues that are caused by the implicit assumption that the deletion requests are independent of the prior machine learning model, and they modify the SISA algorithm to work for so-called \emph{adaptive} sequences of deletion requests. 
Indeed, it may well be the case that a user is more likely to request data deletion upon knowledge of model decisions concerning them.
In \citet{golatkar2020mixedprivacy}, the authors extend their method considered in \Cref{sec:fisher} to apply to deep neural networks. 
A similar task to the one of unlearning data points is of unlearning classes, which we do not consider in this paper, but is considered by \citet{Golatkar_2019} and by others such as \citet{baumhauer2020machine}. \citet{Wu22PUMA} extend the general methodology of the Influence method \citep{guo2020certified} that we discuss in Section \ref{sec:influence}. 
We note that there has been research in applying unlearning to recommender systems and regression problems, notably by \citet{Schelter20} for item-based collaborative filtering recommender systems, $k$-nearest neighbours, and ridge regression. In addition, \citet{chen2022recommender} extend the ideas of the SISA algorithm to recommender systems. 
Finally, an area not considered in this paper concerns Markov Chain Monte Carlo models, for which the first unlearning methods have recently been developed \citep{fu2022knowledge, Nguyen_2022}.

%Introduce related review papers?
%
%These methods attempt to improve on naive retraining...
%
%- Problem statement - comply with legislation, educe computational overhead. \\
%- What has been lacking in the literature - different datasets used, attempts to stnadradise have not been widely adopted. Several methods share common aspects. Comaprisons of models have not been broad e.g. variety of datasets. \\
%
%- Mention addition of data points to model
%- Mention differntial privacy.
%- mention models have tradeoffs that can be controlled.

\section{Terminology}
\label{preliminaries}

In this section, we give formal definitions of the fundamental concepts in machine unlearning, which appear throughout the literature and the rest of this paper. Due primarily to the varying approaches towards achieving unlearning in the literature, definitions of a machine unlearning algorithm vary substantially. The key aim of this section is to provide a definition of unlearning that unifies these approaches.

Throughout this paper, we let $\mathcal{X}$ denote the feature space and $\mathcal{Y}$ denote the output space. A data point is an element of $\mathcal{Z} = \mathcal{X}\times\mathcal{Y}$, and we let $\mathcal{Z}^*$ denote the space of datasets. Explicitly, a dataset $D\in\mathcal{Z}^*$ is a multiset of data points, allowing for duplicate entries. 

The objective of our learning framework is to learn, from a dataset $D$, a hypothesis function $h:\mathcal{X}\to\mathcal{Y}$, which assigns an output $y=h(x)\in\mathcal{Y}$ to a given input $x\in\mathcal{X}$. A training algorithm can therefore be viewed as a map $\mathcal{A}:\mathcal{Z}^*\to\mathcal{H}$, where $\mathcal{H}$ is the space of all hypothesis functions, whose objective is to minimise a non-negative real-valued loss function $L(h, D)$. All training algorithms we consider are randomised, by which we mean that they admit some degree of randomness, for example through random initialisation of weights or through random selection of training data as in stochastic gradient descent. 

In the setting of unlearning, the initial training data $D$ is fixed with $n\geq 1$ samples and a dimensionality of $p\geq 1$. Due to the randomness of $\mathcal{A}$, the output $\mathcal{A}(D)(x)$ at a point $x\in\mathcal{X}$ can be viewed as a random variable. If $\mathcal{A}$ is parametric, then there exists a space of parameters $\theta\in\mathrm{\Theta}$ of fixed dimension determining the hypothesis function $h_{\theta} = \mathcal{A}(D)$, and in this case we use $\theta$ and $h_{\theta}$ interchangeably to denote our trained model. The goal of an unlearning algorithm is to remove the influence of a subset $D_u\subseteq D$ of $m$ samples from the trained machine learning model $\mathcal{A}(D)$. The rest of this section is dedicated to formalising the notion of machine unlearning.

In the following definition we define an \emph{update mechanism}, which is one of two core components to an unlearning algorithm.

\begin{definition}[Update Mechanism]\label{def:update}
An \emph{update mechanism} is a map
\[
	\mathcal{U}:\mathcal{H}\times\mathcal{Z}^*\times\mathcal{Z}^*\to\mathcal{H},
\]
which takes as input a model $h\in\mathcal{H}$, two datasets $D, D_u\in\mathcal{Z}^*$, and outputs a new model $\mathcal{U}(h, D, D_u)\in\mathcal{H}$.
\end{definition}

\begin{remark} In the case of unlearning, we have $D_u\subseteq D$ and the output of the update mechanism aims to remove as much information about $D_u$ as possible from the model $h$ in a way that we make precise in the rest of this section; in this case we may also use \emph{removal mechanism} as a synonym for update mechanism. Definition \ref{def:update} above gives the minimal inputs needed to define an update mechanism. In reality, update mechanisms typically require additional inputs, which are clearly defined in each specific case in later sections. In the initial application of $\mathcal{U}$ in unlearning, we have $h = \mathcal{A}(D)$, however subsequent applications of $\mathcal{U}$ may be applied to models that are not necessarily the result of applying $\mathcal{A}$, but from some preceding application of $\mathcal{U}$ itself, for example. In all cases considered in this paper, the mechanism $\mathcal{U}$ is randomised in the same manner as randomised training algorithms.
\end{remark}

One way of ensuring the removal of $D_u$ from a trained model is to stipulate that the removal mechanism is equivalent to applying the training algorithm $\mathcal{A}$ on the dataset $D$ with $D_u$ removed. This is so-called \emph{exact removal} which is defined as follows. 

\begin{definition}[Exact Removal Mechanism]\label{def:exact}
An update mechanism $\mathcal{U}$ for a randomised training algorithm $\mathcal{A}$ is said to be an \emph{exact removal mechanism} if and only if it achieves the same model that would have been achieved by retraining from scratch. Formally, $\mathcal{U}$ is exact if and only if, for all $D\in\mathcal{Z}^*$, $D_u\subseteq D$, and $x\in\mathcal{X}$, the random variables $\mathcal{U}(\mathcal{A}(D), D, D_u)(x)$ and $\mathcal{A}(D\setminus D_u)(x)$ admit the same distributions.
\end{definition}

The simplest example of exact removal is to just retrain from scratch on $D\setminus D_u$. This is the \emph{na\"ive removal mechanism} defined as follows.

\begin{definition}[Na\"ive Removal Mechanism]\label{def:naiveremoval}
Given a randomised training algorithm $\mathcal{A}$, the \emph{na\"ive removal mechanism} $\mathcal{U}^*$ is the update mechanism defined by
\[
	(\mathcal{A}(D), D, D_u)\mapsto \mathcal{A}(D\setminus D_u),
\]
where $D_u\subseteq D$.
\end{definition}

Whilst na\"ive removal is a simple method to implement and achieve unlearning, it can be computationally expensive and time consuming. Some of the key developments in the unlearning literature involve the development of alternative exact removal mechanisms to na\"ive removal, as we shall see in Section \ref{sec:exact_methods}.

Exact removal is not the only option, however, and \citet{Ginart_2019} proposed mechanisms that achieve removal through alternative means, which are so-called \emph{approximate removal mechanisms}. The probability distribution of these mechanisms no longer match that of na\"ive removal, and we must therefore resort to alternative means of guaranteeing that information on $D_u$ has been removed from the model $h$ -- this guarantee is known as the \emph{certifiability} of the update mechanism. For example, \citet{guo2020certified} guarantee this by proving statistical indistinguishability between the update mechanism and na\"ive unlearning, as follows.

\begin{definition}[$\epsilon$-certifiability, \citet{guo2020certified}]\label{def:epscertified} 
Let $\epsilon > 0$ be given, we say that an update mechanism $\mathcal{U}$ is an \emph{$\epsilon$-certified removal mechanism} with respect to the randomised training procedure $\mathcal{A}$ if for all datasets $D\in\mathcal{Z}^*$, removal subsets $D_u\subseteq D$, inputs $x\in\mathcal{X}$ and output subsets $\mathcal{T}\subseteq\mathcal{Y}$, we have
\begin{equation}
	e^{-\epsilon}\leq \frac{\Pr[\mathcal{U}(\mathcal{A}(D), D, D_u)(x)\in\mathcal{T}]}{\Pr[\mathcal{A}(D\setminus D_u)(x)\in\mathcal{T}]}\leq e^{\epsilon}. \label{eq:epscertified}
\end{equation}
\end{definition}
Intuitively, $\epsilon$-certified removal gives a bound of $e^{\epsilon}$ on the probability that one can distinguish between the updated model and the model obtained through na\"{i}ve removal. 
The notion of $\epsilon$-certified can be weakened slightly to give $(\epsilon, \delta)$-certified removal as follows.

\begin{definition}[$(\epsilon, \delta)$-certifiability, \citet{guo2020certified}]\label{def:epsdeltacertified} 
Let $\epsilon, \delta > 0$ be given. 
We say that an update mechanism $\mathcal{U}$ is an \emph{$(\epsilon, \delta)$-certified removal mechanism} with respect to the randomised training procedure $\mathcal{A}$ if for all datasets $\mathcal{D}\in\mathcal{Z}$, removal subsets $D_u\subseteq D$, inputs $x\in\mathcal{X}$, and output subsets $\mathcal{T}\subseteq \mathcal{Y}$, we have
\begin{align}
	\begin{split}
	\Pr[\mathcal{U}(\mathcal{A}(D), D, D_u)(x)\in\mathcal{T}] &\leq e^{\epsilon}\Pr[\mathcal{A}(D\setminus D_u)(x)\in\mathcal{T}] + \delta, \\
	\Pr[\mathcal{A}(D\setminus D_u)(x)\in\mathcal{T}] &\leq e^{\epsilon}\Pr[\mathcal{U}(\mathcal{A}(D), D, D_u)(x)\in\mathcal{T}] + \delta.
	\end{split}\label{eq:epsdeltacertified}
\end{align}
\end{definition}

The above discussion motivates a definition of unlearning that consists of two components, an update mechanism along with a guarantee of certifiability of this mechanism. However, the varying methods for showing certifiability in the literature make it difficult to consolidate this notion into a standard definition. This is perhaps partly due to a lack of clarity in current regulations concerning the application of unlearning methods in practice, which makes it difficult to argue for the use of certain certifiability guarantees over others. To circumvent this, we stipulate only that the update mechanism be accompanied by a verifiable set of theoretical guarantees and empirical results concerning the mechanism's ability to remove information about $D_u$ from $\mathcal{A}(D)$. The required nature and degree of certifiability is left up to external judgement by theoreticians, practitioners and auditors. 

We now give a straw man concept for a \emph{theoretical} certificate of unlearning, which should be provided in any theoretical development of an unlearning algorithm. This should be extended to a formalised certificate of unlearning concept that applies also in practice, on the same level of rigour as the ``Proof of Learning" concept from \cite{jia2021proofoflearning}. In practice, this should provide enough information (for example source code) for auditors to recreate any historical removals, and subsequently recompute and verify the certificate of unlearning. The need for such a concept has also been identified in \citet{thudi2021necessity}, who provide their own ``Proof of Unlearning". 

\begin{definition}[Certificate of Unlearning] \label{def:certificate} 
Given a randomised training algorithm $\mathcal{A}$ and update mechanism $\mathcal{U}$, a \emph{certificate of unlearning}, $\mathcal{C}_{\mathcal{U}}$ of $\mathcal{U}$, is a set of verifiable claims with which certifiability may be (re-)computed and analysed externally. 
The claims should show that, for any dataset $D$ and subset $D_u\subseteq D$, the model $\mathcal{U}(\mathcal{A}(D), D, D_u)$ contains a sufficiently small amount of information about $D_u$. The precise sense of ``sufficiently small" depends on the particular certifiability claim, which could, for example, be a proof of exactness (Definition \ref{def:exact}) or $\epsilon$-certifiability (Definition \ref{def:epscertified}).
\end{definition}

\begin{definition}[Machine Unlearning Algorithm]\label{def:unlearning}
Given a randomised training algorithm $\mathcal{A}$, a machine unlearning algorithm is a pair $(\mathcal{U}, \mathcal{C}_{\mathcal{U}})$ consisting of an update mechanism $\mathcal{U}$ (Definition \ref{def:update}) and a certificate of unlearning $\mathcal{C}_{\mathcal{U}}$ (Definition \ref{def:certificate}).
\end{definition}

\begin{remark} 
As we shall see later, some of the unlearning algorithms provide only empirical results for the certificate of unlearning. 
While empirical results and measures are a helpful addition to the certificate of unlearning, given the privacy-sensitive nature of many applications of unlearning, they are not enough alone.
The addition of empirical results to the certificate of unlearning is most useful in the case of approximate removal, where removal of data is only guaranteed to an approximate degree.
\end{remark}

For the rest of this section, we fix a randomised training algorithm $\mathcal{A}$ and drop the dependence on $\mathcal{A}$ from notation. We give the baseline example of an unlearning algorithm in this framework.

\begin{definition}[Na\"ive Unlearning]\label{ex:naive}
The na\"ive unlearning algorithm is $(\mathcal{U}^*, \mathcal{C}_{\mathcal{U}^*})$, where $\mathcal{U}^*$ is the na\"ive removal mechanism (Definition \ref{def:naiveremoval}). The certificate of unlearning in this case contains the trivial fact that $\mathcal{U}^*$ is exact.
\end{definition}

\begin{definition}[Exact and Approximate Unlearning Algorithms]\label{def:naive_exact_approx}
An \emph{exact unlearning algorithm} is a pair $(\mathcal{U}, \mathcal{C}_{\mathcal{U}})$, where $\mathcal{U}$ is an exact unlearning mechanism (Definition \ref{def:exact}); in this case $\mathcal{C}_{
\mathcal{U}}$ includes a proof that $\mathcal{U}$ is an exact removal mechanism. An \emph{approximate unlearning algorithm} is an unlearning algorithm that is not exact. 
\end{definition}

While we have mostly formulated the process of machine unlearning above for a one-time batch removal, in reality unlearning may be required to occur over a sequence of removals, as in the case where a user requests the deletion of their data. As a result, analysis of an unlearning algorithm over arbitrary sequences of removals is helpful. The simplest case of sequences of single removals is considered by \citet{Neel_2020}, whose method we detail later in Section \ref{sec:d2d}. They introduce the notion of strong, weak and $(\alpha, \beta)$-accurate unlearning algorithms, as follows. Let $\{\mathbf{z}_i\mid 0\leq i\leq m\}\subseteq D$ be a sequence of data points to be unlearned, and let
\begin{equation}
	D_0 = D,\hspace{10pt} D_i = D_{i-1}\setminus\{\mathbf{z}_{i-1}\},\hspace{10pt} h_0 = \mathcal{A}(D_0),\hspace{10pt}h_i = \mathcal{U}(h_{i-1}, D_{i-1}, \mathbf{z}_{i-1}), \hspace{20pt} 1\leq i\leq m, \label{sequencenotation}
\end{equation}
be the sequence of datasets and unlearned models, respectively, occurring in the unlearning process. The $(\alpha, \beta)$-accuracy provides a theoretical guarantee of the obtained parameters of an unlearning algorithm over the length of the sequence, whereas strong vs. weak characterises algorithms based on their unlearning speed over the length of the sequence. 

\begin{definition}[$(\alpha, \beta)$-accuracy, \citet{Neel_2020}]\label{def:abacc} 
Let $h_i^* = \mathcal{A}(D_{i-1}\setminus\{\mathbf{z}_{i-1}\})$ be the result of na\"ive removal on $h_{i-1}$ for $i=1, \dots, m$. Given $\alpha>0$ and, $0<\beta < 1$, we say that ($\mathcal{U}$, $\mathcal{C}_{\mathcal{U}})$ is \emph{$(\alpha, \beta)$-accurate} if, for all $0\leq i\leq m$, we have
\begin{equation}
	\Pr\left[L(h_i, D_i) - L(h_i^*, D_i) > \alpha\right] < \beta. \label{eq:abacc}
\end{equation}
In other words, the probability that the unlearned loss and the na\"ive retrained loss differing by more than $\alpha$ is at most $\beta$.
\end{definition}

\begin{definition}[Strong and Weak Unlearning, \citep{Neel_2020}]\label{def:strongweak} 
Let $C_i$ denote the time taken to perform the $i$th removal. 
Assume that $(\mathcal{U}, \mathcal{C}_{\mathcal{U}})$ is $(\alpha, \beta)$-accurate and that $\alpha$ and $\beta$ are independent of $m$.
\begin{enumerate}[(i)]
	\item The unlearning algorithm $(\mathcal{U}, \mathcal{C}_{\mathcal{U}})$ is said to be \emph{strong} if the update time is at most logarithmic in the length of the update sequence. That is, for all $1\leq i\leq m$ we have 
\[
	C_i/C_1 = \mathcal{O}(\log(i)).
\]
	\item We say that $(\mathcal{U}, \mathcal{C}_{\mathcal{U}})$ is \emph{weak} if the update time is polynomial in the length of the update sequence. That is, for all $1\leq i\leq m$ we have 
\[
	C_i/C_1 = \Omega(\poly(i)).
\]
\end{enumerate}
\end{definition}

To finish this section, we define two concepts which are closely related to the theory of machine unlearning.
Differential privacy inspired the definition of $(\epsilon, \delta)$-certifiability in \citet{guo2020certified}, and we can see the similarities of the definition of differential privacy below to that of Definition \ref{def:epsdeltacertified}.
Intuitively, differential privacy bounds the effect on the output of a query of changing a singleton's data; as a result, it constrains the amount of information an attacker could extract about an individual in a dataset.

\begin{definition}[Differential Privacy, \citet{Dwork_2014}]\label{def:diffpriv} 
We say that a randomised training algorithm $\mathcal{A}$ is \emph{$(\epsilon, \delta)$-differentially private} if for all pairs of datasets $D_1$ and $D_2$ that differ by a singleton's data, for all inputs $x\in\mathcal{X}$ and output subsets $\mathcal{T}\subseteq\mathcal{Y}$, we have
\begin{equation}
	\Pr[\mathcal{A}(D_1)(x) \in \mathcal{T}] \leq e^{\epsilon}\Pr[\mathcal{A}(D_2)(x) \in \mathcal{T}] + \delta. \label{eq:diffpriv}
\end{equation}
\end{definition}

%where $\delta$ is the probability of information being leaked and $\epsilon$ is the maximum distance between queries on datasets $x$ and $y$, i.e. the amount of information that can be inferred about an individual. The larger the value of $\epsilon$, the weaker the constraint on the dataset.

\begin{definition}[Catastrophic Forgetting, \citet{Kirkpatrick_2017}] \label{def:catforget} 
Catastrophic forgetting is the rapid decline in predictive ability of a model on previously learned tasks when fine-tuned for a new task.
\end{definition}

\begin{remark} 
Catastrophic forgetting is unsuitable for machine unlearning as information on the original training set can still be extracted from the weights of the fine-tuned model, as discussed in \citet{Golatkar_2019}.
\end{remark}

\section{Evaluation Approaches for Unlearning Algorithms}
\label{sec:evaluation}

Comprehensive evaluation of a machine unlearning algorithm $(\mathcal{U}, \mathcal{C}_{\mathcal{U}})$ is achieved through consideration of four key properties -- \emph{efficiency}, \emph{effectiveness}, \emph{consistency}, and \emph{certifiability} (\textbf{EECC}) -- defined as follows.

\begin{definition}[Efficiency]\label{def:efficiency} 
The \emph{efficiency} of $(\mathcal{U}, \mathcal{C}_{\mathcal{U}})$ is the relative speed-up of the removal mechanism $\mathcal{U}$ over the na\"ive removal mechanism $\mathcal{U}^*$.
\end{definition}

\begin{definition}[Effectiveness]\label{def:effectiveness} 
The \emph{effectiveness} of $(\mathcal{U}, \mathcal{C}_{\mathcal{U}})$ is the test set performance of the unlearned machine learning model, $\mathcal{U}(\mathcal{A}(D), D, D_u)$, in comparison to the na\"{i}ve retrained model's test set performance.
\end{definition}

\begin{definition}[Consistency]\label{def:consistency} 
The \emph{consistency} of $(\mathcal{U}, \mathcal{C}_{\mathcal{U}})$ is a measure of similarity between the unlearned model, $\mathcal{U}(\mathcal{A}(D), D, D_u)$, and the na\"{i}ve retrained model. 
\end{definition}

\begin{definition}[Certifiability]\label{def:certifiability} 
The \emph{certifiability} of $(\mathcal{U}, \mathcal{C}_{\mathcal{U}})$ is the ability of the removal mechanism $\mathcal{U}$ to remove the information of $D_u$ from the unlearned model, $\mathcal{U}(\mathcal{A}(D), D, D_u)$. Precise measures and guarantees of this ability form the contents of $\mathcal{C}_{\mathcal{U}}$.
\end{definition}

Typically in the literature, these four measures are evaluated empirically on real-world and synthetic datasets, often accompanied by additional theoretical analysis.
In experiments, a range of datasets are chosen to reflect varying task complexities.

Within each method, empirical performances for EECC can be tuned through certain parameters.
There are pairwise trade-offs between efficiency, effectiveness, and certifiability as demonstrated in \citet{Mahadevan_2021}.
In addition, there are trade-offs between efficiency, effectiveness and consistency.
Each method contains parameters that directly control the efficiency of the method, which are called \emph{efficiency parameters}; these can be used to improve efficiency at the cost of effectiveness and certifiability.
Approximate methods may also have \emph{certifiability parameters} that directly control the certifiability of the method.

We detail EECC in the following sections, giving empirical measures for them.  
\emph{Performance metric} refers to a metric that is used to evaluate the machine learning models, for example, accuracy.

\subsection{Efficiency}
\label{sec:efficiency}

Efficiency of an unlearning algorithm $(\mathcal{U}, \mathcal{C}_{\mathcal{U}})$ is empirically measured by the ratio of the time taken to obtain the unlearned model $h^{\mathcal{U}} = \mathcal{U}(\mathcal{A}(D), D, D_u)$ to the time taken to obtain the na\"{i}ve retrained model $h^* = \mathcal{U}^*(\mathcal{A}(D), D, D_u)$, defined as follows:
\begin{equation}
	\speed(h^{\mathcal{U}}) := \frac{\text{time taken to obtain $h^*$}}{\text{time taken to obtain $h^{\mathcal{U}}$}}. \label{speedup}
\end{equation}
In practice, this is usually achieved by measuring time taken to obtain $h^*$ and $h^{\mathcal{U}}$ directly, averaged over a number of deletion requests. 
In \citet{Brophy_2021}, efficiency is measured by the number of samples that can be unlearned in the time it takes to unlearn \emph{one} instance by na\"{i}ve retraining; 
this measure is equivalent to the definition in (\ref{speedup}) above. 
Time complexities are often provided as theoretical guarantees, which highlight the role of certain parameters as efficiency parameters.

\subsection{Effectiveness}
\label{sec:effectiveness}

Effectiveness can be empirically measured by comparing the test set performance of the unlearned model to the test set performance of the na\"{i}ve retrained model. Let $y_{\pred}^*$ and $y_{\pred}^{\mathcal{U}}$ be predictions from the na\"{i}ve retrained and unlearned model on a test set $y_{\test}$.
Given a real-valued performance metric $\mathcal{M}$, let $\mathcal{M}_{\test}^* = \mathcal{M}(y_{\pred}^*, y_{\test})$ and $\mathcal{M}_{\test}^{\mathcal{U}} = \mathcal{M}(y_{\pred}^{\mathcal{U}}, y_{\test})$ denote the test set performance, respectively, of the na\"{i}ve retrained model and the unlearned model. 
An example of a measure of the effectiveness of a machine unlearning algorithm $(\mathcal{U}, \mathcal{C}_{\mathcal{U}})$ is then given by the absolute difference of these two quantities:
\begin{equation}
	\eff(h^{\mathcal{U}}) := |\mathcal{M}_{\test}^{\mathcal{U}} -  \mathcal{M}_{\test}^*|. \label{effectiveness}
\end{equation}
The smaller $\eff$ is, the closer in performance the unlearned model is to the na\"{i}ve retrained model, and so smaller values of $\eff$ are preferable. 

\subsection{Consistency}
\label{sec:consistency}

Consistency is a measure of how close an unlearned model is to the na\"{i}ve retrained model; 
that is, a measure of whether the unlearning algorithm is producing the ideal machine learning model as its output. 
Exact unlearning methods are guaranteed a high level of consistency. In contrast to efficiency and effectiveness, the literature we consider contains a number of measures for consistency, which differ in their usage and applicabilities. In this section we include three different measures.

Consistency of a machine unlearning algorithm $(\mathcal{U}, \mathcal{C}_{\mathcal{U}})$ applied to \emph{parametric models} can be measured by the distance between the unlearned and na\"ive retrained parameters:
\begin{equation}
	 \paramcon(h_{\theta}^{\mathcal{U}}) := \|\theta^{\mathcal{U}} - \theta^*\|_2, \label{paramconsistency}
\end{equation}
where $\theta^{\mathcal{U}}$ and $\theta^*$ are the parameters of $h^{\mathcal{U}}$ and $h^*$ respectively, and $\|\cdot\|_2$ denotes the $\ell^2$-norm. This is used, for example, in \citet{Wu_2020}. Lower values of $\paramcon$ are preferable as they indicate that the parameters of the unlearned model are close to those of the na\"ive retrained model.

In \citet{he2021deepobliviate}, a measure of consistency that is applicable to classification models is considered. Consistency is given as the proportion of predictions that agree between na\"{i}ve retrained and unlearned:
\begin{equation}
	\outputcon(h^{\mathcal{U}}) := \frac{100}{n_{\test}}\sum_{i=1}^{n_{\test}}1_{y_{\pred, i}^* = y_{\pred, i}^{\mathcal{U}}}, \label{outputconsistency}
\end{equation}
where $y_{\pred, i}^*$ and $y_{\pred, i}^{\mathcal{U}}$ are the na\"ive retrained and unlearned model predictions, respectively, on a test set containing $n_{\test}$ samples.
Higher values of $\outputcon$ are preferable, since they indicate that test-set predictions of the unlearned model are similar to those of the na\"ive retrained model.
Note that lower consistency may be acceptable if effectiveness and certifiability remain high. 
In this case, the correct predictions of the unlearned model differ but high performance on a test set is maintained; 
consistency, particularly $\outputcon$, is still useful in this situation as a measure of divergence from the na\"{i}ve retrained model's predictions but it does not necessarily indicate an undesirable machine unlearning algorithm. The $\outputcon$ measure is limited to classification models only, however analogues for regression tasks may be considered. 

In \citet{Golatkar_2019}, the Kullback-Leibler (KL) divergence is minimised as part of the loss function. 
KL-divergence measures the similarity of two probability distributions $P$ and $Q$ as follows 
\[
	\kl(P\| Q) := \mathbb{E}_{X\sim P}\left[\log\left(\frac{p(X)}{q(X)}\right)\right],
\]
where $p$ and $q$, respectively, are the probability functions (either mass or density) of $P$ and $Q$.
By Gibbs' inequality, we have $\kl(P\|Q)\in[0, \infty)$. Smaller values of $\kl$ give higher similarity between $P$ and $Q$. This gives a measure of consistency of an unlearning algorithm applied by measuring the KL-divergence of the unlearned model distribution to that of the na\"ive retrained model:
\begin{equation}
	\conkl(h^{\mathcal{U}}) := \kl(\pr[h^{\mathcal{U}}(x)]\ \| \pr[h^*(x)]). \label{kl} 
\end{equation}
Lower values of $\conkl$ are preferable. The KL-divergence may also be used to measure similarity of probability outputs between the unlearned model and the na\"ive retrained model, when applied to classification tasks.

\subsection{Certifiability}
\label{sec:certifiability}

Certifiability is assurance that the unlearned model has removed information about the deleted data point and that an unlearning request has been fully complied to with respect to regulations.
An important part of this involves quantifying and minimising vulnerability to inference attacks on the deleted data, which involve the identification of the deleted data by external attackers. 
Exact unlearning methods are guaranteed to remove the influence of the data point to be forgotten, however they may still be vulnerable to inference attacks based around the deleted data points. In such a case, the na\"{i}ve retrained model is also equally vulnerable to such inference attacks. To minimise this vulnerability, changes may need to be made to the training procedure, such as the addition of noise to gradient updates.

Consistency, from the previous section, is something of a precursor to certifiability, and certifiability was developed in order to address the insufficiency of consistency for guaranteeing deletion, see, for example, \citet[p. 2]{guo2020certified}. 
We keep them separate because some definitions and measures of certifiability do not guarantee consistency and, moreover, the notion of certifiability is slightly removed from that of consistency on a fundamental level. 
Consistency is a correctness guarantee, whereas certifiability is a security guarantee. 

Measures and guarantees of certifiability differ greatly across the literature. 
Often, only theoretical guarantees are given and proven, such as $\epsilon$- and $(\epsilon, \delta)$-certifiability (Definitions \ref{def:epscertified}, \ref{def:epsdeltacertified}) in \citet{guo2020certified}. 
Empirical measures are given in \citet{he2021deepobliviate}, \citet{Mahadevan_2021}, and \citet{Golatkar_2019}, which we outline in the rest of this section.

The accuracy of the na\"{i}ve retrained model on the removed data $D_u$ quantifies the amount of information that the model can be expected to have on the removed data, having never seen this data. This might be quite high if the remaining data contains similar data, for example. This accuracy is therefore a baseline with which the accuracy of the unlearned model on $D_u$ can be compared, as done in \citet{Mahadevan_2021}.
Here, we extend this to a general performance metric $\mathcal{M}$. Let $\mathcal{M}_u^* := \mathcal{M}(y_u^*, y_u)$ and $\mathcal{M}_u^{\mathcal{U}} := \mathcal{M}(y_u^{\mathcal{U}}, y_u)$ denote the  performance on $D_u$, respectively, of the na\"{i}ve retrained model and the unlearned model after unlearning $D_u$. Certifiability is measured by:
\begin{equation}
	\certdis(h^{\mathcal{U}}) :=  \frac{|\mathcal{M}_u^{\mathcal{U}} - \mathcal{M}_u^*|}{|\mathcal{M}_u^{\mathcal{U}}| + |\mathcal{M}_u^*|}\cdot 100. \label{certdis}
\end{equation}
The right-hand side of Equation (\ref{certdis}) is called the \emph{symmetric absolute percentage error} -- $\sape(\mathcal{M}_u^{\mathcal{U}}, \mathcal{M}_u^*)$ -- of $\mathcal{M}_u^*$ and $\mathcal{M}_u^{\mathcal{U}}$. It is used here to penalise deviations more when $\mathcal{M}_u^*$ is small, guided by the intuition that $\mathcal{M}_u^* = 0\%$ and $\mathcal{M}_u^{\mathcal{U}} = 10\%$, say, represents perhaps a more serious breach of information than $\mathcal{M}_u^* = 80\%$ and $\mathcal{M}_u^{\mathcal{U}} = 90\%$. 

Large values of $\certdis$ imply that the unlearned model contains information about the deleted data that it should not, so this measure can indicate degradation of certifiability between models, or of a single model over time. However, since $\certdis$ measures only the similarity  of performance on the deleted set, small values do not necessarily imply high levels of certifiability such as those seen in theoretical guarantees. For example, if the na\"{i}ve unlearned model has high predictive performance on $D_u$ and $\certdis$ is small, then the success rate at predicting on $D_u$ is similar for all three models -- original, na\"{i}ve, and unlearned. In this case, small $\certdis$ does not distinguish whether the unlearned model has forgotten the data, since it will achieve high performance on $D_u$ whether $D_u$ is present in the training data or not. To address this issue, \citet{he2021deepobliviate} adapt and apply the backdoor verification experiment from \citet{sommer2020probabilistic}. In this experiment, $D_u$ is ``augmented'' and a specific target label is associated to this augmented data; such augmented data and corresponding label is called backdoor data. The result is that a model is likely to predict the correct specific backdoor label only if it has been trained on $D_u$. Hence the na\"{i}ve retrained model is guaranteed to have poor predictive performance on $D_u$, which allows $\certdis$ to be a more meaningful measure of certifiability.

In \citet{Golatkar_2019}, the Shannon Mutual Information is used as a measure of certifiability. This measures the amount of information that two random variables $X$ and $Y$ share and is defined by
\begin{equation}\label{eq:smi}
	I(X; Y) := \mathbb{E}_x[\kl(P_{Y\mid X} \| P_Y)],
\end{equation}
where $P_{Y\mid X}$ is the conditional probability distribution of $Y$ with respect to $X$, and $P_Y$ is the probability distribution of $Y$.
By treating the removal subset $D_u$ as a random variable (through, for example, random selection of removal points), the mutual information that $D_u$ and the unlearned model share is given by $I(D_u; \mathcal{U}(\mathcal{A}(D), D, D_u))$. It is shown \citep[Proposition 1, Lemma 1]{Golatkar_2019} that this mutual information is bounded from above by a specific KL-divergence that is more readily computable. 
In particular, once $D_u$ is chosen and fixed, the consistency measure $\conkl$ of (\ref{kl}) provides an empirical upper bound on the mutual information.
This procedure is specific to their method and is not applicable in general.

\section{Exact Unlearning Algorithms}
\label{sec:exact_methods}

In this section we describe two exact unlearning algorithms. First we describe the SISA algorithm, which is one of the most broadly applicable methods we include, before considering a highly specialised algorithm called DaRE, which applies only to decision-tree based machine learning models.

% -------------------------------------------------------------------------------------------------------------------
\subsection{SISA}
\label{sec:sisa}

The SISA algorithm is an exact unlearning algorithm introduced in \citet{Bourtoule_2021}, borrowing aspects from both ensemble learning and distributed training to efficiently unlearn.
This is achieved by a reorganisation of the training dataset, known as sharding and slicing, which reduces the time needed to retrain from scratch with the specified data removed. 

The full SISA algorithm is applicable to any machine learning model that has been trained incrementally, for example, via gradient descent. The loss function for such models need not be strongly convex, and \citet{Bourtoule_2021} apply SISA successfully to deep neural networks (DNN) in a variety of architectures. SISA without slicing is applicable to all machine learning models, including decision trees.

\subsubsection{Methodology.}

\begin{figure}
	\centering
	\includegraphics[width=0.5\textwidth, angle=270]{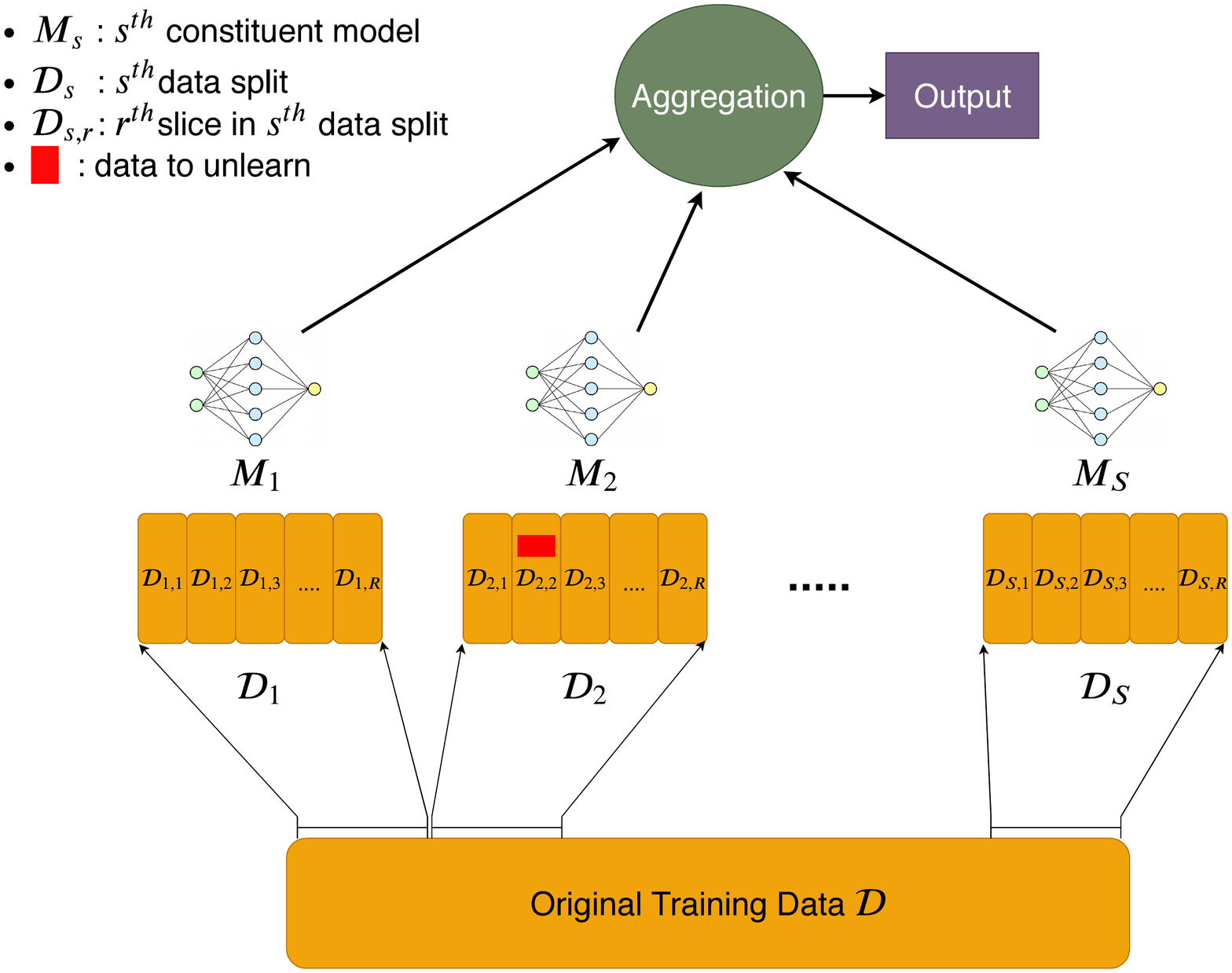}
	\caption{SISA training, taken from \citet{Bourtoule_2021}. 
	The training data is split into $S$ shards, which are further split into $R$ slices. $S$ independent models are trained incrementally on the slices, and predictions of these models are aggregated to form a final output. 
The data to unlearn is highlighted in red in this diagram. To unlearn this data point, only $M_2$ needs to be retrained, and this process starts from slice $\mathcal{D}_{2, 2}$. In our notation (Algorithms \ref{alg:SISA_training}, \ref{alg:sisaunlearn}) model $M_2$ is denoted by $h_{2}$, and the saved intermediary model state $\tilde{h}_{2, j}$, for each $1\leq j\leq R$, correspond to the model trained on the slices $\bigcup_{r\leq j} \mathcal{D}_{2, r}$.}
	\label{fig:sisa_diagram}
\end{figure}

\citet{Bourtoule_2021} define the SISA training process to consist of four key steps -- Sharded, Isolated, Sliced, and Aggregated -- which can be seen in \Cref{fig:sisa_diagram} and which are described below. The resultant machine learning model is an ensemble of weak learners. Pseudocode for the SISA training algorithm can be found in Algorithm \ref{alg:SISA_training}.

\begin{enumerate}[(i)]
\item\textbf{Sharded.}
The original training dataset is separated into approximately equal-sized shards, with each training data point contained in exactly one shard. 
This contrasts with other ensemble modelling techniques, in which training data points can be present in many of the learners that make up the ensemble model. 
\item\textbf{Isolation.}
Each of the shards is trained in isolation from the other shards, restricting the influence of each data point to a single shard.  
\item\textbf{Slicing.} 
Each of the shards are sub-divided into slices, which are presented to the algorithm incrementally as training proceeds. 
The trained model states are saved after each slice.
\item\textbf{Aggregation.} 
To form the final model prediction for a data point, the predictions of each sharded model are aggregated, which can be done through a variety of methods.
\end{enumerate}
Whenever a removal request for a single data point comes in, only the model trained on the shard containing the particular data point needs to be retrained and, moreover, retraining needs only begin from the slice containing the data point. As a result, the expected retraining time is faster compared to na\"{i}ve retraining; the exact speed-up depends on the number of shards and slices used, as discussed below. Algorithms \ref{alg:SISA_training} and \ref{alg:sisaunlearn} contain pseudocode for the training and unlearning procedures, which assume no prior knowledge of the distribution of deletion requests, choosing shards and slices randomly. Prior knowledge of the distribution of deletion requests allows improvement of the SISA method and is discussed in \citet[Section VIII]{Bourtoule_2021}. In Algorithms \ref{alg:SISA_training}, \ref{alg:sisaunlearn}, the function $\textsc{Train}(D\mid \tilde{h})$ produces a model by initialising at the model state $\tilde{h}$ and training on the dataset $D$.

Several approaches to aggregating during the inference phase are possible. 
A simple label-based majority vote involves each constituent model contributing equally to the ensemble model's final decision. 
Other options include averaging the output vectors of class or continuous predictions from each individual model. 
The latter method was not found to have a noticeable effect on accuracy for simple datasets, compared to the label-based majority vote, however mean vector aggregation has higher accuracy for the more complex ImageNet task \citep[Figure 13, 14]{Bourtoule_2021}.

\begin{algorithm}
\caption{Initial training with SISA.}
\label{alg:SISA_training}
\begin{algorithmic}[1]
\Statex \textbf{Input:} training data $D$, number of shards $S$, number of slices $R$, number of epochs for each slice $\mathbf{e}$.
\Statex \textbf{Output:} ensemble of models $h = (h_1, \dots, h_S)$ and intermediary model states $\tilde{h} = (\{\tilde{h}_{i, 0}, \dots, \tilde{h}_{i, R}\})_{i=1}^S$.
\Procedure{SisaTrain}{$D$; $S$, $R$, $\mathbf{e}$}
	\State split the data randomly into shards  $D_1, \dots, D_S$ and save shard indices for each data point
	\State split each shard $D_i$ randomly into $R$ slices $D_{i, 1}, \dots D_{i, R}$ and save slice indices for each data point
	\State randomly initialise $(\tilde{h}_{1, 0}, \dots, \tilde{h}_{S, 0})$
    \For {$i=1; i \leq S; i++$}
    	\For {$j=1; j \leq R; j++$} 
    		\State $h_{i, j}\leftarrow\textsc{Train}\left(D_{i, 1}\cup \cdots \cup D_{i, j}\mid \tilde{h}_{i, j-1}\right)$ for $e_j$ epochs
		\State save model state $\tilde{h}_{i, j}$ of model $h_{i, j}$
    	\EndFor
	\State $h_i\leftarrow h_{i, R}$ 
    \EndFor
\State\Return $h =  (h_1, \dots, h_S)$, $\tilde{h}^{\mathcal{U}} = (\{\tilde{h}_{i, 0}, \dots, \tilde{h}_{i, R}\})_{i=1}^S$.
\EndProcedure
\end{algorithmic}
\end{algorithm}

\begin{algorithm}
\caption{SISA removal mechanism.}
\label{alg:sisaunlearn}
\begin{algorithmic}[1]
\Statex \textbf{Input}: ensemble of models $h =  (h_1, \dots, h_S)$, saved intermediary states $\tilde{h} = (\{\tilde{h}_{i, 0}, \dots, \tilde{h}_{i, R}\})_{i=1}^S$, data $D$ with saved shard-slice indices, removal data $D_u$, number of shards $S$, number of slices $R$, epochs $\mathbf{e}$.
\Statex \textbf{Output}: unlearned model $h^{\mathcal{U}} = (h_1^{\mathcal{U}}, \dots, h_S^{\mathcal{U}})$ and intermediary model states $\tilde{h}^{\mathcal{U}} = (\{\tilde{h}_{i, 0}^{\mathcal{U}}, \dots, \tilde{h}_{i, R}^{\mathcal{U}}\})_{i=1}^S$.
\Procedure{SisaUnlearn}{$h$, $\tilde{h}$, $D$, $D_u$; $S$, $R$, $\mathbf{e}$}
	\For {$i=1; i \leq S; i++$}
	\State initialise model $h_{i, R}^{\mathcal{U}}\leftarrow h_i$ and states $\{\tilde{h}_{i, 0}^{\mathcal{U}}, \dots, \tilde{h}_{i, R}^{\mathcal{U}}\}\leftarrow \{\tilde{h}_{i, 0}, \dots, \tilde{h}_{i, R}\}$
    	\If {$\exists z\in D_u$ with shard index $i$}
    		\State $r_i\leftarrow$ find minimal slice index for all $z\in D_u$ that have shard index $i$
		\For{$j=r_i; j\leq R; j++$}
			\State $D_{i, j}'\leftarrow D_{i, j}\setminus (D_u\cap D_{i, j})$
			\State $h_{i, j}^{\mathcal{U}}\leftarrow\textsc{Train}\left(D_{i, 1}\cup\cdots\cup D_{i, r_i-1}\cup D_{i, r_i}'\cup\cdots \cup D_{i, j}'\mid \tilde{h}_{i, j-1}^{\mathcal{U}}\right)$ for $e_j$ epochs
			\State save model state $\tilde{h}_{i, j}^{\mathcal{U}}$ for $h_{i, j}^{\mathcal{U}}$
		\EndFor
    	\EndIf
	\State $h_i^{\mathcal{U}}\leftarrow h_{i, R}^{\mathcal{U}}$
    \EndFor
\State\Return $h^{\mathcal{U}} = (h_1^{\mathcal{U}}, \dots, h_S^{\mathcal{U}})_i$, $\tilde{h}^U = (\{\tilde{h}_{i, 0}^{\mathcal{U}}, \dots, \tilde{h}_{i, R}^{\mathcal{U}}\})_{i=1}^S$
\EndProcedure
\end{algorithmic}
\end{algorithm}

\subsubsection{Performance.}
The datasets used for the evaluation of the SISA algorithm include the Purchase \citep{Sakar_2018}, \textsc{svhn} \citep{Netzer_2011}, and the ImageNet \citep{Deng_2009} datasets. 

\paragraph{Efficiency.}
With 20 shards and 50 slices per shard, 8 unlearning requests on the Purchase dataset resulted in an $\speed$ of 4.63$\times$, with 18 unlearning requests on \textsc{svhn} resulting in a $\speed$ of 2.45$\times$.
The removal of 39 data points from the more complex ImageNet learning task gave a $\speed$ of 1.36$\times$.

The number of shards, $S$, is an efficiency parameter. As discussed in \citet[Section VII.A]{Bourtoule_2021}, increasing the number of shards increases the efficiency of SISA, but will degrade the predictive performance of the resultant machine learning model compared to a lower number of shards. For the simpler learning tasks tested (\textsc{svhn} and Purchase datasets) increasing $S > 20$ entails a more noticeable drop in accuracy. Increasing the number of slices in each shard, $R$, reduces the retraining time but this does not degrade accuracy provided that the epochs in training are carefully chosen \citep[Sections V, VII]{Bourtoule_2021}. However, an increase in $R$ does come at increased storage costs due to the increased number of saved model states. As discussed in \emph{ibid.}, (Appendix C), the efficiency-storage trade-off of $R$ may be preferable to the efficiency-effectiveness trade-off of $S$. \citet{Bourtoule_2021} derive a maximum theoretical efficiency of $\frac{(R+1)S}{2}\times$.

\paragraph{Effectiveness.}
The experiments on simple tasks -- Purchase and \textsc{svhn} -- outlined above resulted in $\eff$ values of less than 2\%. The more complex ImageNet removal observed a larger $\eff$ of 18.76\%. 

\paragraph{Consistency.} 
Consistency is guaranteed by the exact nature of the SISA removal mechanism.

\paragraph{Certifiability.}
\citet{Bourtoule_2021} argue that as the SISA mechanism inherently retrains without the removed data, this shows that SISA is exact. 
The certificate of unlearning is therefore provided by exactness in this case.
However, if an attacker has access to the model before and after an unlearning request is made, then the SISA method is vulnerable to having information about the deleted data points inferred, as discussed in \citet{Golatkar_2019}.

% -------------------------------------------------------------------------------------------------------------------
\subsection{DaRE Forests}

\label{sec:dare}

In `Machine Unlearning for Random Forests', \citet{Brophy_2021} introduce an exact (\emph{ibid.}, Theorem 3.1) unlearning algorithm that is specific to decision-tree and random-forest based machine learning models for binary classification. 
This is done through the development of \emph{\textbf{Da}ta \textbf{R}emoval-\textbf{E}nabled} (\textbf{DaRE}) \emph{trees}, and the ensemble of these to form \emph{DaRE Forests} (\textbf{DaRE RF}). 
Through the use of strategic thresholding at decision nodes for continuous attributes, high-level random nodes, and caching certain statistics at all nodes, DaRE trees enable efficient removal of training instances. 

\subsubsection{Methodology.}

The DaRE forest is an ensemble consisting of DaRE trees in which each tree is trained independently on a copy of the training data, which differs from the data bootstrapping of traditional random forests. 
Aside from this, the DaRE forest ensembles in the same way as a random forest, in particular a random subset of $\tilde{p}$ features are considered at each split.

In the rest of this section, we focus on DaRE trees, which must be initially trained in the DaRE methodology of Algorithm \ref{alg:dare_train}.
As in regular decision trees, DaRE trees are trained recursively by selecting, at most nodes, an attribute and threshold that optimises a \emph{split criterion} (Gini index in \citet{Brophy_2021}). 
They differ from regular decision trees in three key ways as follows.

\begin{enumerate}[(i)]

\item\textbf{Random nodes.}
The top $d_{\rmax}$ levels of nodes in a DaRE tree are random nodes, where $d_{\rmax}$ is an integer hyperparameter.
Random nodes are defined as nodes for which both an attribute $a$ and a threshold in the attribute range $[a_{\min}, a_{\max})$ are chosen uniformly at random. 

\item\textbf{Threshold sampling.} During training and deletion, DaRE trees randomly sample $k$ \emph{valid} thresholds at any node that is neither a random node nor a leaf.
These are thresholds that lie between two adjacent data points with opposite labels. 
Doing so reduces the amount of statistics one needs to store at each node and speeds up computation. 

\item\textbf{Statistics caching.}
At each node, the number of data points $|D_{\node}|$ and $|D_{\node, 1}|$ are stored and updated, where $D_{\node}$ is the input dataset to the node and $D_{\node, 1}\subseteq D_{\node}$ is the subset of positive instances. 
For each of the $k$ candidate valid thresholds $v$, various additional statistics are stored and updated. The exact form of these depends on the type of node \citep[see][Appendix A.6]{Brophy_2021}. In each case these statistics are sufficient to recompute the split criterion scores and to determine the validity of the current thresholds. 
At leaf nodes, pointers to the training data at that leaf are stored.
As a result, the removal mechanism is able to recall training data from the stored leaf instances, meaning that training data is not required as an explicit input to the mechanism. 

\end{enumerate}

\begin{algorithm}
\caption{$\textsc{DareTrain}(D, 0; d_{\rmax}, k)$ trains a single DaRE tree \citep{Brophy_2021}.}
\label{alg:dare_train}
\begin{algorithmic}[1]
\Statex\textbf{Input:} data $D_{\node}$, depth $d$, random node depth $d_{\rmax}$, threshold candidate size $k$.
\Statex\textbf{Output:} trained subtree rooted at a level-$d$ node.
\Procedure{DareTrain}{$D_{\node}$, $d$; $d_{\rmax}$, $k$}      
\If{stopping criteria reached}
	\State $\node\leftarrow$ \textsc{LeafNode}()
	\State save instance counts $|D_{\node}|$, $|D_1|$
	\State save leaf-instance pointers$(node, D_{\node})$
	\State compute leaf value$(node)$
\Else
	\If{$d < d_{\rmax}$}
		\State $\node\leftarrow$ \textsc{RandomNode}()
		\State save instance counts $|D_{\node}|$, $|D_{\node, 1}|$
		\State $a\leftarrow$ randomly sample attribute$(D_{\node})$
		\State $v\leftarrow$ randomly sample threshold $\in[a_{\min}, a_{\max})$
		\State save threshold statistics$(\node, D_{\node}, a, v)$
	\Else
		\State $\node\leftarrow$ \textsc{GreedyNode}()
		\State save instance counts $|D_{\node}|$, $|D_{\node, 1}|$
		\State $A\leftarrow$ randomly sample $\tilde{p}$ attributes$(D_{\node})$
		\For{$a\in A$}
			\State $C\leftarrow$ get valid thresholds$(D_{\node}, a)$
			\State $V\leftarrow$ randomly sample $k$ valid thresholds$(C)$
			\For{$v\in V$}
				\State save threshold statistics$(\node, D_{\node}, a, v)$
			\EndFor
			\State $scores\leftarrow$ compute split scores$(\node)$
			\State select optimal split$(\node, scores)$
		\EndFor
	\EndIf
	\State $D_{\text{left}}, D_{\text{right}}\leftarrow$ split on selected threshold$(\node, D_{\node})$
	\State $\node.\text{left} = \textsc{DareTrain}(D_{\text{left}}, d+1; d_{\rmax}, k)$
	\State $\node.\text{right} = \textsc{DareTrain}(D_{\text{right}}, d+1; d_{\rmax}, k)$
\EndIf
\State \Return $\node$
\label{alg:dare_training}
\EndProcedure
\end{algorithmic}
\end{algorithm}

\begin{algorithm}
\caption{Deleting a training instance from a DaRE tree, \citep{Brophy_2021}.}
\label{alg:dare_delete}
\begin{algorithmic}[1]
\Require{start at the root node.}
\Statex\textbf{Input:} $\node$, data point to remove $\mathbf{z}$, depth $d$, random node depth $d_{\rmax}$, threshold candidate size $k$.
\Statex\textbf{Output:} retrained subtree rooted at node.
\Procedure{DareUnlearn}{$\node$, $\mathbf{z}$, $d$; $d_{\rmax}$, $k$}
\State update instance counts $|D_{\node}|$, $|D_{\node, 1}|$
\If{node is a \textsc{LeafNode}}
	\State remove $\mathbf{z}$ from leaf-instance pointers$(\node, \mathbf{z})$
	\State recompute leaf value$(\node)$
	\State remove $\mathbf{z}$ from database and return
\Else
	\State update decision node statistics$(\node, \mathbf{z})$
	\If{node is a \textsc{RandomNode}}
		\If{$\node.selectedThreshold$ is invalid}
			\State $D_{\node}\leftarrow$ get data from the set of leaf instances$(\node)\setminus \{\mathbf{z}\}$
			\If{$\node.selectedAttribute(a)$ is not constant}
				\State $v\leftarrow$ resample threshold $\in[a_{\min}, a_{\max})$
				\State $D_{\node, \ell}, D_{\node, r}\leftarrow$ split on new threshold$(\node, D_{\node}, a, v)$
				\State $\node.\ell\leftarrow$ \textsc{DareTrain}$(D_{\node, \ell}, d+1; d_{\rmax}, k)$
				\State $\node.r\leftarrow\textsc{DareTrain}(D_{\node, r}, d+1; d_{\rmax}, k)$
			\Else
				\State $\node\leftarrow$ \textsc{DareTrain}$(D_{\node}, d; d_{\rmax}, k)$
			\EndIf
			\State remove $\mathbf{z}$ from database and return
		\EndIf
	\Else
		\If{$\exists$ invalid attributes or thresholds}
			\State $D_{\node}\leftarrow$ get data from the set of leaf instances$(\node)\setminus\{ \mathbf{z}\}$
			\State resample invalid attributes and thresholds$(\node, D_{\node})$
		\EndIf
		\State $scores\leftarrow$ recompute split scores$(\node)$
		\State $a, v\leftarrow$ select optimal split$(\node, scores)$
		\If{optimal split has changed}
			\State $D_{\node.\text{left}}, D_{\node.\text{right}}\leftarrow$ split on new threshold$(\node, D_{\node}, a, v)$
			\State $\node.\text{left}\leftarrow$ \textsc{DareTrain}$(D_{\node.\text{left}}, d+1; d_{\rmax}, k)$
			\State $\node.\text{right}\leftarrow\textsc{DareTrain}(D_{\node.\text{right}}, d+1; d_{\rmax}, k)$
			\State remove $\mathbf{z}$ from database and return
		\EndIf
	\EndIf
	\If{$x_{a}\leq v$}
		\State\textsc{DareUnlearn}$(\node.\text{left}, \mathbf{z}, d+1; d_{\rmax}, k)$
	\Else
		\State\textsc{DareUnlearn}$(\node.\text{right}, \mathbf{z}, d+1; d_{\rmax}, k)$
	\EndIf
\EndIf
\label{alg:dare_unlearn}
\EndProcedure
\end{algorithmic}
\end{algorithm}

Deletion of a data point $\mathbf{z}$ is given in Algorithm \ref{alg:dare_delete}. 
Only those subtrees that have been trained on $\mathbf{z}$ are considered. 
At each decision node that is neither random nor a leaf, statistics are recalculated and the split criterion is updated for each attribute-threshold pair.
If a different optimal threshold is found then the subtree rooted at this node is retrained. 
At the relevant leaf node, the pointer to $\mathbf{z}$ is deleted. 
Random nodes are rarely retrained, only if the randomly chosen threshold no longer lies within the attribute range $[a_{\min}, a_{\max})$.

The DaRE RF inherits the following hyperparameters of random forests: $\tilde{p}$ is the number of random attributes to consider at each split; $d_{\max}$ is the maximum depth for each tree; $T$ is the number of trees in the forest. 
DaRE RFs have two additional hyperparameters: $k$ is the number of random valid thresholds to sample at each split; $d_{\rmax}$ is the number of top layers that are used for random nodes.

\subsubsection{Performance.}

The authors evaluate DaRE RFs on 13 real-world binary classification datasets and one synthetic dataset via efficiency and effectiveness; the proof of exactness, \citep[Theorem 3.1]{Brophy_2021}, covers certifiability. 

For the experiments considered by \citet{Brophy_2021}, hyperparameters are chosen or tuned as follows. The parameter $\tilde{p}$ is set at $\lfloor\sqrt{p}\rfloor$. 
The parameters $d_{\max}$, $T$, and $k$ are tuned using 5-fold cross-validation, with $d_{\rmax} = 0$ fixed, to maximise the relevant performance metric for the task at hand, using the Gini index as the split criterion. 
The parameter $d_{\rmax}$ is incremented from $d_{\rmax} = 0$ until the 5-fold cross-validation score breaches four separate absolute error loss tolerances $\{0.1, 0.25, 0.5, 1.0\}$ relative to the DaRE RF with $d_{\rmax} = 0$, yielding four DaRE RFs with varying proportions of random nodes. Larger error tolerances lead to larger values of $d_{\rmax}$ \citep[Appendix B.2, Table 6]{Brophy_2021}.
Performance of all five DaRE RF models -- $d_{\rmax} = 0$ and the four tuned $d_{\rmax}$ -- are considered.

\paragraph{Efficiency.} 
Efficiency is measured as the number of training instances that can be deleted by DaRE RF in the time it takes to delete one instance via na\"{i}ve retraining, which is equivalent to $\speed$ (\ref{speedup}). 
When deleting points at random, the average $\speed$ of DaRE RFs is 2--4 orders of magnitude. 
In the best case, DaRE RF achieves a $\speed$ of 35,856$\times$, achieved on the \textsc{higgs} dataset and with the highest proportion of random nodes; the average $\speed$ across all 14 datasets for the highest $d_{\rmax}$ DaRE RF is 1,272$\times$. 
A worse-case deletion strategy called \emph{worst-of-1000} is also considered, giving average $\speed$ of 1--3 orders of magnitude.

Note that DaRE RF training has the same time complexity, $\mathcal{O}(T\tilde{p}nd_{\max})$, as the training for a traditional random forest \citep[Theorem 3.2]{Brophy_2021}. 
Time complexities for deleting a single instance are given in \emph{ibid.} (Theorem 3.3); the best-case time complexity occurs when the tree structure is unchanged and all attribute thresholds remain valid, which is $\mathcal{O}(\tilde{p}kd_{\max})$, with additional costs occurring when thresholds become invalid and subtrees require retraining.

\paragraph{Effectiveness.} 
Test set performance of each DaRE RF is compared to the DaRE RF with no random nodes. In most cases, $\eff < 1\%$.
Larger proportions of random nodes generally degrades performance, particularly for the Credit Card and \textsc{higgs} datasets, for which we see $\eff\approx 2\%$. 

In \citet[Appendix B.2.]{Brophy_2021}, the predictive performances of the various DaRE RFs are compared to the scikit-learn implementation of random forests. As shown in Table 5, \emph{ibid.}, DaRE RFs with random nodes have worse performance than the standard random forest across all datasets, but in some cases the DaRE RF with no random nodes improves on performance over the scikit-learn random forest.

The level of random nodes in a DaRE RF, $d_{\rmax}$, is an efficiency parameter, with larger values entailing faster unlearning at the cost of predictive performance.
This is demonstrated in \citet[Appendix B.2, Table 6]{Brophy_2021}, in which higher error tolerances lead to higher proportions of random nodes. 
Test set error is shown to increase dramatically once $d_{\rmax}$ gets too large \citep[Figure 2]{Brophy_2021}.
The number of valid thresholds to consider, $k$, is another efficiency parameter. 
Reducing $k$ will increase efficiency, however predictive performance suffers \citep[Figure 3]{Brophy_2021}. 
Predictive performance plateaus after $k = 5$ for the Surgical dataset, which gives an optimal choice of $k =5$ in this case.

\paragraph{Consistency and certifiability.}
Consistency and certifiability of the DaRE algorithm are guaranteed by the exact nature of the removal mechanism, as proven by \citet[Theorem 1.3]{Brophy_2021}. 
The certificate of unlearning for DaRE is therefore provided by exactness.

\section{Approximate Unlearning Algorithms}
\label{sec:approx_methods}

Approximate unlearning algorithms originate from a proposal in the work of \citet{Ginart_2019}. In this section we describe five approximate methods, some of which build upon the ideas of \citet{Ginart_2019}. They are given in chronological order.

% -------------------------------------------------------------------------------------------------------------------
\subsection{Fisher}
\label{sec:fisher}

The Fisher algorithm, introduced by \citet{Golatkar_2019}, applies to parametric models and is an approximate unlearning algorithm that 
updates the parameters of a pre-trained convex model in a time-efficient manner. The method facilitates the 
removal of the information about certain data without retraining the model from scratch, via a strategic mix of Newton correction and noise injection. Precise upper bounds on the amount of information that can be extracted about the removed data can be calculated.

\subsubsection{Methodology.}
\citet{Golatkar_2019} develop a batch unlearning removal mechanism in which a subset of 
the data $D_u\subseteq D$ is unlearned from a trained parametric model. They formalise the definition of 
``complete'' forgetting by introducing the condition that forgetting of information can only be guaranteed if the KL divergence, (\ref{kl}), between the probability distributions of outputs from the unlearned and the na\"ively retrained models, is zero. By \citet[Proposition 1]{Golatkar_2019}, this also ensures that there is zero Shannon Mutual Information,  $I(D_u; \mathcal{U}(h_{\theta}, D, D_u)) = 0$, as defined in \Cref{eq:smi}, between the deleted data $D_u$ and the unlearned model, $\mathcal{U}(h_{\theta}, D, D_u)$. The authors therefore stress on minimising the KL divergence as part of the unlearning objective function. This can be achieved by updating the parameters of the initial model through a single Newton step on the remaining data, or by adding Gaussian noise in the direction of this Newton step, or both \citep[see][Corollary 1]{Golatkar_2019}. Upon Newton correction and noise injection to the weights of the original model, the Fisher training and unlearning method are respectively expressed as:
\begin{align}
\theta :&= \textsc{sgd}(L(\theta_{\init}, D)) + \sigma F^{-\frac{1}{4}}\mathbf{b}, \\
\theta^{\mathcal{U}} :&= \underbrace{\theta-F\textsuperscript{-1}\mathrm{\Delta}_{\rem}}_{\text{Newton step}}+\underbrace{\sigma F\textsuperscript{-1/4} \mathbf{b},}_{\text{noise injection}}
\end{align}
where $\textsc{sgd}$ is the stochastic gradient descent algorithm; $\theta_{\init}$ are the initialised parameters; $\theta^{\mathcal{U}}$ are the unlearned parameters; $\sigma > 0$ is the noise parameter; $F$ is the Fisher Information Matrix \citep[Eq. (8)]{Golatkar_2019}; $\mathbf{b}\sim\mathcal{N}(0, 1)^p$ is Gaussian noise, where recall that $p$ is the dimensionality of the training data; and
\begin{equation}
	\begin{split}
		\mathrm{\Delta}_{\rem} :&= \nabla L(\theta, D\setminus D_u), \\
	\end{split} \label{hessian}
\end{equation}
denotes the gradient of the loss function of $\theta$ on the remaining data.
The Fisher Information matrix is used as an approximation to the Hessian as the Fisher matrix is less expensive to compute. In some cases, such as when the loss function is the log-likelihood (for example, \citet{Mahadevan_2021} who consider linear logistic regression), the Fisher Information Matrix and Hessian coincide.
The algorithm for the deletion of data points in mini-batches from \citep{Mahadevan_2021} using this method is shown in Algorithm \ref{alg:fisher}.

\begin{algorithm}
\caption{Fisher removal mechanism, \citep{Golatkar_2019, Mahadevan_2021}.}
\label{alg:fisher}
\begin{algorithmic}[1]
 	\Statex\textbf{Input}: trained model parameters $\theta$, training dataset $D$, subset of data to be removed $D_u$, noise parameter $\sigma$, mini-batch size $m'$.
	\Statex\textbf{Output}: unlearned model parameters $\theta^{\mathcal{U}}$.
	\Procedure{FisherUnlearn}{$\theta$, $D$, $D_u$; $\sigma$, $m'$}
  \State assign the number of batches $s\leftarrow\lceil \frac{m}{m'}\rceil$ ($m$ is the number of samples in $D_u$)
	\State split $D_u$ into $\emph{s}$ mini-batches {$D^{1}_{u}$, $D^{2}_{u}$,...,$D^{s}_{u}$}, each of size $m'$
  \State initialise $D^{\prime} \leftarrow D$
  \State initialise $\theta^{\mathcal{U}} \leftarrow \theta$
  \For{$i=1; i \leq s;\; i++$}
  		\State $D^{\prime} \leftarrow D^{\prime} \setminus D_u^i$
      \State $\mathrm{\Delta}\leftarrow {\nabla L(\theta^{\mathcal{U}},D^{\prime})}$
      \State $F \leftarrow$ compute Fisher Information Matrix of $L$ and $D'$, \cite[Eq. (8)]{Golatkar_2019}
      \State ${\theta^{\mathcal{U}}} \leftarrow {\theta^{\mathcal{U}}} - F\textsuperscript{-1}\Delta$
  \If{$\sigma > 0$}
	\State draw $\mathbf{b}\sim\mathcal{N}(0, 1)^p$
      \State ${\theta^{\mathcal{U}}} \leftarrow {\theta^{\mathcal{U}}} + \sigma F\textsuperscript{-1/4}\mathbf{b}$
  \EndIf    
  \EndFor
  \State \Return $\theta^{\mathcal{U}}$
\EndProcedure
\end{algorithmic}
\end{algorithm}

\subsubsection{Performance.}

\paragraph{Efficiency.}
Apart from the original paper, \citet{Mahadevan_2021} evaluate the performance of the Fisher method across datasets varying in dimensionality and size. They show that the Fisher method is a moderately efficient unlearning method across the range of dimensionalities (ranging from 28 to 3072) and sizes (from $\sim 10^4$ to $\sim 10^6$) of the datasets. Controlling for the noise parameter, $\sigma = 1$, and using only one mini-batch, they found that the method offered an $\speed$ of up to 50$\times$ for the low-dimensional datasets while only around 1.8$\times$ for the high-dimensional datasets. The large difference in the algorithm's performance across different dimensionalities is due to the difference in the cost of computing the inverse Hessian matrix, which is much faster to calculate when the dimensionality is low. 

The mini-batch size $m'$ used by \citet{Mahadevan_2021} is an efficiency parameter. It controls a trade-off between efficiency and both effectiveness and certifiability (\emph{ibid.} (Figures 3, 4)), since a smaller number of batches (large mini-batch size) ensures that the Fisher matrix is calculated a fewer number of times at the cost of fewer Newton steps.
When measuring the effectiveness-efficiency trade-off, the authors find that the method suffers a huge loss in efficiency, especially in the high dimensional setting where the $\speed$ of the method reduced to as low as 0.4$\times$. The authors attribute this poor efficiency to the computational effort it takes to inject noise post update, which takes longer when the dimensionality is high. In the low dimensionality setting, $\speed$ reduced to 9$\times$, and high effectiveness was maintained. 

\paragraph{Effectiveness.} In \citet[Table 1]{Golatkar_2019}, effectiveness of the Fisher method when unlearning a subset of 100 images from the Lacuna-10 and \textsc{cifar}-10 datasets is approximately $4-5\%$ in each case, measured with $\eff$. This translates to approximately $2-3\%$ in \emph{percentage error}, which is the measure used by \citet[Eq. (19)]{Mahadevan_2021}, who achieve percentage error of $\leq 1\%$ across all 6 tested datasets. The better effectiveness observed in \citet{Mahadevan_2021} results from the use of mini-batches in the update stage.

\paragraph{Consistency.} Consistency is given in \citet[Proposition 2]{Golatkar_2019} by providing a computable upper bound for the KL-divergence measure $\conkl$ of (\ref{kl}). This bound is 3.3 for the Lacuna-10 dataset and 33.4 for \textsc{cifar}-10. Consistency is not measured in \citet{Mahadevan_2021}.

\paragraph{Certifiability.}
As shown by \citet{Golatkar_2019}, the KL-divergence measure (\ref{kl}) is an upper bound for the Shannon Mutual Information. Therefore, the Shannon Mutual Information between $D_u$ and the unlearned model is at most 3.3 kNATs for the Lacuna-10 dataset and 33.4 kNATs for \textsc{cifar}-10. In \citet[Figure 3]{Mahadevan_2021}, certifiability is shown using $\certdis$ for different values of the noise parameter $\sigma$. For small volumes and $\sigma = 0$, $\certdis$ mostly has values between 0\% and 1\%, however a notable large value of $\certdis\approx 10\%$ is observed for the \textsc{higgs} dataset. Values of $\certdis$ decrease for $\sigma = 1$, entailing better certifiability. Note, however, that the increase in $\sigma$ degrades the effectiveness of the method since large amount of 
noise yields less optimal parameters and lower accuracy for such models.

% -------------------------------------------------------------------------------------------------------------------
\subsection{Influence}

\label{sec:influence}

The Influence algorithm, introduced by \citet{guo2020certified}, is an approximate unlearning 
algorithm that applies to parametric models and leverages the ML influence theory \citep{koh2020understanding} to unlearn. Specifically, the 
method performs unlearning by quantifying the influence of the deleted data on the original 
model's parameters and applying a Newton update to these parameters to remove the 
identified influence of the deleted data. Despite employing a Newton update removal mechanism 
for unlearning similar to the Fisher method (Section \ref{sec:fisher}), this method takes a 
different approach to noise injection in order to guarantee certifiability. More 
specifically, the noise in this method is added to the gradient loss at 
every step of the initial training of the model, unlike in the Fisher method in which noise is added 
at the end of the training and unlearning procedures. The authors justify this choice by 
stressing that the direction of the gradient residual at the time of parameter update may 
reveal information about the deleted data, which can be prevented by adding the noise at the 
time of training.

\subsubsection{Methodology.}
The noise-injected objective function used for training this method is expressed as:
\begin{equation}
L_{\sigma}(\theta, D) := L(\theta, D) + \dfrac{{\sigma}{\textbf{b}
^T}{\theta}}{|D|},
\label{eqn:noisyloss}
\end{equation}
where ${\sigma}$ is the noise control parameter, $\theta$ is the vector of model parameters,
${\textbf{b}}\sim \mathcal{N}(0, 1)^p$ is Gaussian noise.

Training for Influence optimises the noisy loss (\ref{eqn:noisyloss}). Note that noise is added at every step of the optimisation procedure, in contrast to Fisher which adds noise only at the end. Moreover loss is only added during training time and not during the removal mechanism (where the original loss function $L(\theta, D)$ is used). To unlearn, a single Newton update on the model parameters $\theta$ is performed as follows:
\begin{equation}
\theta^{\mathcal{U}} := \theta + {H}^{-1}{\mathrm{\Delta}_u},
\label{eqn:influenceupdate}
\end{equation}
where
\begin{align}
\mathrm{\Delta}_u :&= \nabla L(\theta, D_u), \\
H :&= \nabla^2L(\theta, D\setminus D_u),
\end{align}
denote, respectively, the gradient of the loss function on the deleted data and the Hessian of the loss function (without noise) on the remaining data.
The so-called influence function $H^{-1}\mathrm{\Delta}_u$ gives the direction, $H$, and magnitude, $\mathrm{\Delta}_u$, of the step required to remove the influence of the deleted data from $\theta$. The Influence removal mechanism is shown in Algorithm \ref{alg:influence_alg}.

\begin{algorithm}[H]
\caption{Influence removal mechanism, \citep{Mahadevan_2021}.}
\label{alg:influence_alg}
\begin{algorithmic}[1]
	\Statex\textbf{Input}: trained model parameters $\theta$, original train dataset $D$, subset of data to be deleted $D_u$, mini-batch size $m'$.
	\Statex\textbf{Output}: unlearned model parameters $\theta^{\mathcal{U}}$.
	\Procedure{InfluenceUnlearn}{$\theta$, $D$, $D_u$; $m'$}
  \State assign the number of batches $s\leftarrow\lceil \frac{m}{m'}\rceil$ ($m$ is the number of samples in $D_u$)
	\State split $D_u$ into $\emph{s}$ mini-batches {$D^{1}_u$, $D^{2}_u$,...,$D^{s}_u$}, each of size $m'$
	\State initialise $D'\leftarrow D$
	\State initialise $\theta^{\mathcal{U}}\leftarrow\theta$
	\For{$i = 1; i\leq s; i++$}
		\State $D'\leftarrow D'\setminus D_u^i$
		\State $\mathrm{\Delta}_{m'}\leftarrow\nabla L\left(\theta^{\mathcal{U}}, D_u^i\right)$
		\State $H\leftarrow \nabla^2 L\left(\theta^{\mathcal{U}}, D'\right)$
		\State $\theta^{\mathcal{U}}\leftarrow\theta^{\mathcal{U}} + H^{-1}\mathrm{\Delta}_{m'}$
  \EndFor
\State\Return $\theta^{\mathcal{U}}$
\EndProcedure
\end{algorithmic}
\end{algorithm}

\subsubsection{Performance.}

\paragraph{Efficiency.} \citet{Mahadevan_2021} evaluate the performance of the Influence method across datasets varying in dimensionality and sizes of the training datasets, as described in Section \ref{sec:fisher}. Controlling for the noise parameter, $\sigma = 1$, they found that the method offered a $\speed$ of up to 200$\times$ for the low-dimensional datasets while only around 5$\times$ for the high-dimensional datasets on bulk removals. The large difference in the method's performance across different dimensionalities is due to the difference in the cost of computing the inverse Hessian matrix, which is much faster to calculate when the dimensionality is low. 

As in the Fisher method, Section \ref{sec:fisher}, the mini-batch size $m'$ acts as an efficiency parameter, which is because smaller mini-batch sizes lead to a larger number of calculations of the Hessian of the loss function. The effect of the trade-off between efficiency and both effectiveness and certifiability is discussed in \citet[Figures 3, 4]{Mahadevan_2021}.

\paragraph{Effectiveness.} \citet{Mahadevan_2021} found that despite being highly efficient, the Influence method did not compromise effectiveness too much. For example, Influence is no more than $1\%$ less effective than Fisher when evaluated using the \emph{percentage error} (\emph{ibid.}, Eq. (19)).

\paragraph{Consistency.} In \citet{guo2020certified}, consistency is measured by considering the $\ell^2$-norm of the gradient residual $\|\nabla L(\theta^{\mathcal{U}}, D\setminus D_u)\|_2$. For the true optimiser $\min_{\theta}L(\theta, D\setminus D_u)$ this gradient residual is zero, hence the value $\|\nabla L(\theta^{\mathcal{U}}, D\setminus D_u)\|_2$ measures the error of the parameters $\theta^{\mathcal{U}}$ in approximating the true optimiser. Upper bounds for this gradient residual are provided in \emph{ibid.} (Theorems 1, 2; Corollary 1, 2), which give non-data-dependent and data-dependent bounds for single and batch deletion. These are empirically verified on \textsc{mnist} in \emph{ibid.} (Fig. 2). Consistency is not measured in \citet{Mahadevan_2021}.

\paragraph{Certifiability.} For a desired $\epsilon > 0$, the Influence method is shown to be $\epsilon$- and $(\epsilon, \delta)$-certified for an $\epsilon$-dependent choice of $\sigma$ given in \citet[Theorem 3]{guo2020certified}. In \citet{Mahadevan_2021}, certifiability is measured using $\certdis$. 
Increasing $\sigma$ allows for smaller $\epsilon$ and gives larger $\certdis$ values, thereby increases the certifiability of the model, however, as shown in \citet[Figure 1]{guo2020certified} and \citet[Figure 5]{Mahadevan_2021}, this comes at the cost of a degradation in effectiveness. It is interesting to note that in \citet{guo2020certified}, test accuracy is relatively stable until a certain value of $\sigma$, after which it experiences dramatic decline.

% -------------------------------------------------------------------------------------------------------------------
\subsection{DeltaGrad}

\label{sec:deltagrad}

The DeltaGrad algorithm is first described in \citet{Wu_2020}. 
It is an approximate unlearning algorithm that uses information cached from the initial training process to more efficiently compute model parameters after data points have been removed. DeltaGrad is only applicable to parametric machine learning models trained using gradient descent or mini-batch stochastic gradient descent, with loss functions that are strongly convex and smooth \citep[Section 2.3]{Wu_2020}.

\subsubsection{Methodology.}
Throughout this section, $L(\theta) = L(\theta, D)$ denotes the empirical loss function on the full training dataset.
An initial model is trained using a gradient descent algorithm, giving a vector of trained model parameters $\theta$. 
At each step of training, $t$, the model parameters $\{\theta_0, \theta_1,...\theta_t\}$ and the gradients of the loss function $\{\nabla L(\theta_0), \nabla L(\theta_1),...\nabla L(\theta_t)\}$ are saved. 

In contrast to the Fisher and Influence methods of Sections \ref{sec:fisher}, \ref{sec:influence} which unlearn by performing gradient descent steps starting from the original optimised model parameters, DeltaGrad performs gradient descent from the same initial parameters of the original training procedure and recalculates all gradient descent steps. The time to retrain is reduced from that of na\"ive retraining by calculating only some of the gradient descent steps exactly and approximating all other steps. Specifically, the first $j_0$ steps and every $T_0$ step thereafter are calculated explicitly, where $j_0, T_0 > 0$ are integer hyperparameters. 

Gradient steps, when deleting $m$ data points with indices in $M$ (i.e., $D_u = \{\mathbf{z}_i\mid i\in M\}\subseteq D$), are approximated by first rewriting the gradient descent update formula as follows \citep[Eq. (2)]{Wu_2020}:
\begin{equation}
\label{leaverout}
	\theta^{\mathcal{U}}_{t+1}\leftarrow \theta_t^{\mathcal{U}} - \frac{\eta_t}{n-m}\left[n\nabla L(\theta_t^{\mathcal{U}}) - \sum_{i\in M}\nabla L_i(\theta_t^{\mathcal{U}})\right],
\end{equation}
where $n$ is the number of data points in $D$.
The value $n\nabla L(\theta_t^{\mathcal{U}})$ is approximated, with the rest of the terms in (\ref{leaverout}) calculated explicitly; under the assumption $m\ll n$ the latter explicit calculation is comparatively inexpensive.
The quantity $n\nabla L(\theta_t^{\mathcal{U}})$ can be expressed, using the Cauchy mean-value theorem \citep[Eq. (3)]{Wu_2020}, in terms of an integrated Hessian $H_t$. 
The L-BFGS algorithm\footnote{\href{https://link.springer.com/article/10.1007/BF01589116}{https://link.springer.com/article/10.1007/BF01589116}} then approximates the vector product $H_t\cdot\mathbf{v}$ as a quasi-Hessian product $B_t\cdot \mathbf{v}$. Put together, this allows the gradient step of (\ref{leaverout}), to be approximated as:
\begin{equation}\label{eqn:deltagrad_approx}
	\theta^{\mathcal{U}}_{t+1} \leftarrow \theta^{\mathcal{U}}_{t} - \frac{\eta_t}{n-m} \left[ n(\nabla L(\theta_t) + B_t \cdot (\theta^{\mathcal{U}}_t - \theta_t)) - \displaystyle\sum_{i \in M} \nabla L_i(\theta^{\mathcal{U}}_t) \right].
\end{equation}
To compute the quasi-Hessian $B_t$, L-BFGS uses $k$ previously stored parameter and gradient differences, $\theta_j^{\mathcal{U}} - \theta_j$ and $\nabla L(\theta_j^{\mathcal{U}}) - \nabla L(\theta_j)$, where $k > 0$ is an integer hyperparameter. The two differences are stored only during the explicitly computed gradient steps, during the first $j_0$ steps and every $T_0$ step thereafter.

DeltaGrad is extended to mini-batch stochastic gradient descent in \citet[Section 3]{Wu_2020}, with a randomly sampled mini-batch $\mathcal{B}$ of size $B$. Algorithm 4, \emph{ibid.}, further extends DeltaGrad to DNNs by checking whether a loss function is convex and smooth locally. 
In this case, the arrays of historical parameter updates $\mathrm{\Delta} G$ and gradients $\mathrm{\Delta} W$ are only saved when local convexity holds and an exact GD update is used otherwise.
This extension of the method adds an additional computational cost to unlearning.

The work of \citet[Equation 16]{Mahadevan_2021} expands this algorithm slightly by following the removal mechanism, described in Algorithm \ref{alg:deltagrad}, with the injection of a Gaussian noise vector controlled by a noise parameter, $\sigma$, in order to control the trade-off between effectiveness and certifiability.

\begin{algorithm}[tbh]
\caption{DeltaGrad removal mechanism, \citep{Wu_2020}}\label{alg:deltagrad}
\begin{algorithmic}[1]
\Statex \textbf{Input:} model training parameters $\boldsymbol{\theta} := \{\theta_0, \theta_1,...\theta_t\}$, training data $D$,  indices of removed training samples $M$, stored training gradients $\nabla L(\boldsymbol{\theta}) := \{\nabla L(\theta_0), \nabla L(\theta_1),...\nabla L(\theta_t)\}$, period $T_0$, total iteration number $T$, history size $k$, burn-in iteration number $j_0$, learning rate $\eta_t$.
\Statex \textbf{Output:} unlearned model parameters $\theta^{\mathcal{U}} = \theta_T^{\mathcal{U}}$.
\Procedure{DeltaGradUnlearn}{$\boldsymbol{\theta}$, $D$, $M$; $\nabla L(\boldsymbol{\theta})$, $T_0$, $T$, $k$, $j_0$, $\eta_t$}
\State initialise $\theta^{\mathcal{U}}_{0} \leftarrow \theta_{0}$
\State initialise an array $\mathrm{\Delta} G\leftarrow []$
\State initialise an array $\mathrm{\Delta} \mathrm{\Theta}\leftarrow []$
\State $\ell\leftarrow 0$
	\For{$t=0; t \leq T; ++$}
		\If{$[(t_0 - j_0)\pmod{T_0} ==0]~or~t \leq j_0$}
    		\State compute $\nabla L(\theta^{\mathcal{U}}_t)$ exactly
    		\State compute $\nabla L(\theta^{\mathcal{U}}_t) - \nabla L(\theta_t)$, using the cached gradient $\nabla L(\theta_t)$
    		\State $\mathrm{\Delta} G[\ell] \leftarrow \nabla L (\theta^{\mathcal{U}}_t) - \nabla L (\theta_t)$
    		\State $\mathrm{\Delta} \mathrm{\Theta}[\ell] \leftarrow \theta^{\mathcal{U}}_t - \theta_t$
    		\State $\ell \leftarrow \ell+1$
    		\State compute $\theta^{\mathcal{U}}_{t+1}$ by using exact GD update
		\Else
			\State $B_{j_k}\leftarrow\textsc{L-BFGS}(\mathrm{\Delta} G[-k:], \mathrm{\Delta} \mathrm{\Theta}[-k:])$
			\State $\nabla L(\theta^{\mathcal{U}}_t)\leftarrow\nabla L(\theta^{\mathcal{U}}_t) + B_{j_k}(\theta^{\mathcal{U}}_t-\theta_t)$ approximate the gradient
    		\State compute $\theta^{\mathcal{U}}_{t+1}$ via the modified gradient formula, Eq. (\ref{eqn:deltagrad_approx}), using approximated $\nabla L(\theta^{\mathcal{U}}_t)$
		\EndIf
	\EndFor
\State\Return $\theta^{\mathcal{U}}_t$
\EndProcedure
\end{algorithmic}
\end{algorithm}

\subsubsection{Performance.}
\citet{Wu_2020} evaluate DeltaGrad applied to logistic regression for the \textsc{mnist}, Covtype, \textsc{higgs}, and \textsc{rcv1} datasets. With the inclusion of noise injection, \citet{Mahadevan_2021} compared the performance of DeltaGrad to Fisher and Influence methods of Sections \ref{sec:fisher}, \ref{sec:influence}.

\paragraph{Efficiency.}
\citet{Wu_2020} show that $\speed$ values for DeltaGrad are 2.5$\times$, 2$\times$, 1.8$\times$ and 6.5$\times$ on the \textsc{mnist}, Covtype, \textsc{higgs}, and \textsc{rcv1} datasets, respectively. It is noted by this work that the theoretical efficiencies that would be expected from this method are not fully achieved, suggesting that a significant portion of this discrepancy originates in the L-BFGS computation of the Hessian projection. \citet[Fig. 3(b)]{Mahadevan_2021} shows that the DeltaGrad method with noise injection is actually slower than na\"ive retraining for the low and medium dimensionality datasets (Covtype, \textsc{higgs}, \textsc{mnist}), but gives $\speed$ of $\sim$2.0-2.5$\times$ for the high-dimensionality datasets of \textsc{cifar2} and Epsilon. 

Under the five assumptions of \citet[Section 2.3]{Wu_2020}, DeltaGrad guarantees convergence at the rate $o\left(\frac{m}{n}\right)$ (\emph{ibid.}, Theorems 1, 2). This shows that DeltaGrad is strong (Definition \ref{def:strongweak}), since we have $m = 1$ for sequential deletions, yielding a convergence rate, $o\left(\frac{1}{n}\right)$, that is independent of the position in the sequence.

\paragraph{Effectiveness.}
In their experiments, \citet[Table 1]{Wu_2020} obtain the same predictive performances, within error, after DeltaGrad as for after na\"ive retraining on the \textsc{mnist}, Covtype, \textsc{higgs} and \textsc{rcv1} datasets, indicating very high effectiveness. In contrast, however, \citet[Fig. 4]{Mahadevan_2021} show that the effectiveness of DeltaGrad is generally lower than the Fisher and Influence algorithms. It can also be seen there that the hyperparameter $T_0$ is an efficiency parameter, decreasing it will increase efficiency due to fewer explicit gradient steps, but causes a slight drop in effectiveness.

\paragraph{Consistency.}
In \citet[Table 2]{Wu_2020}, consistency is measured via $\paramcon$, which measures the $\ell^2$ distances between the unlearned parameters and the na\"ive retrained parameters. Deletion from the \textsc{mnist} dataset showed the largest parameter difference of $2 \times 10^{-4}$, with deletion from the \textsc{rcv1} dataset showing the smallest difference in parameters of $3.5 \times 10^{-6}$. 
Theorem 1, \emph{ibid.}, provides a theoretical guarantee of consistency.

\paragraph{Certifiability.}
While certifiability is not discussed in the original paper \citep{Wu_2020}, the certifiability of the model is measured in \citet{Mahadevan_2021} using $\certdis$. The hyperparameter $\sigma$ acts as a certifiability parameter, increasing it will increase certifiability by providing lower $\certdis$ scores, however this generally comes at the cost of efficiency and effectiveness, as demonstrated by \citet[Figs. 3, 5]{Mahadevan_2021}. It is also shown there that DeltaGrad generally achieves comparable values of $\certdis$ to the Fisher and Influence methods across all datasets when $\sigma = 1$.

% -------------------------------------------------------------------------------------------------------------------
\subsection{Descent-to-Delete}

\label{sec:d2d}

In `Descent-to-Delete: Gradient-Based Methods for Machine Unlearning', \citet{Neel_2020} propose several gradient-based unlearning algorithms for convex parametric models.
These algorithms apply to arbitrary length sequences of addition as well as removal requests in which each request gets handled immediately before the next comes in, however we focus only on removal requests here.
The requests are handled using gradient descent steps starting from the optimum of the original model for the first request and from the state after the most recent update for future requests.
They prove that under certain assumptions these approaches offer constant or logarithmic run-time in relation to the number of requests.
In addition, under stronger assumptions they also offer $(\epsilon, \delta)$-certifiability (Definition \ref{def:epsdeltacertified}) of the full internal model state.
Their paper, however, focuses on theoretical results and does not offer any empirical evidence of performance.

Their work also contributes several terminologies to the unlearning literature.
Firstly, they distinguish perfect and imperfect unlearning algorithms. 
Perfect unlearning algorithms require $(\epsilon, \delta)$-certifiability of the full internal state (i.e., all optimised parameters obtained throughout the deletion sequence) of the algorithm versus the less stringent requirement of $(\epsilon, \delta)$-certifiability of only the final outputs \cite[Definition 2.4]{Neel_2020}.
Secondly, their paper introduces the concept of a strong unlearning algorithm (Definition \ref{def:strongweak}) that updates in at most logarithmic run-time in relation to the sequence of delete requests.

%%%%%%%%%%%%%%%%%%%%%%%%%%%%%%%%%%%%%%%%%%%%%%%%%%%%%%%%%%%%%%%%%%%%%%%%%%%%%%
%%%%%%%%%%%%%%%%%%%%%%%%%%%%%%%%%%%%%%%%%%%%%%%%%%%%%%%%%%%%%%%%%%%%%%%%%%%%%%
\subsubsection{Methodology}

\paragraph{Basic perturbed gradient descent.}
Their first algorithm unlearns a sequence of data points $\{\mathbf{z}_i\}_{i\geq 0}$ from models that have strongly convex loss functions and applies projected gradient descent updates upon each deletion request using the projection function $\operatorname{Proj}_{\mathrm{\Theta}}(\theta) = \operatorname{arg\, min}_{\theta^{\prime} \in \mathrm{\Theta}} \| \theta - \theta^{\prime}\|_{2}$. This is followed by a small perturbation of the model's parameters to guarantee $(\epsilon, \delta)$-certifiability.
This noise removes the ability of an attacker, with access to the model before and after unlearning, to identify data points that were forgotten. After each unlearning procedure, updated datasets $D_i$ and unlearned models $h_{\theta_i}$ form the basis of the input for the following procedure in the sequence of requests, as specified in (\ref{sequencenotation}).
If perfect deletion is not required, the unlearned parameters of the $i$th update, $\hat{\theta}_{i}$, in \Cref{alg:d2d-bpgd} may be used as the input for the subsequent update instead of the published output parameters $\tilde{\theta}_{i}$.
They also extend this algorithm to the non-strongly convex case by introducing a variant with $\ell^2$ regularization.

A key hyperparameter of these algorithms is the number of gradient descent updates to take following a deletion request, $T_i$.
In the regularized model, the number of updates must be balanced against the amount of noise added to the final parameters.
Another key hyperparameter is the variance of the noise, $\sigma$, which is calculated using estimates of the Lipschitz constant and the smoothness of the convex loss function in \citet[Theorems 3.1, 3.2]{Neel_2020} for the unregularized and regularized versions of the algorithm, respectively.

\begin{algorithm}[ht]
\caption{Perfect $i$th unlearning for basic perturbed gradient descent, \citep{Neel_2020}.}
\label{alg:d2d-bpgd}
\begin{algorithmic}[1]
\Statex \textbf{Input:} published model parameters $\tilde{\theta}_{i-1}$, dataset $D_{i-1}$, update data point $\mathbf{z}_{i-1}$, number of iterations $T_i$, noise parameter $\sigma > 0$.
\Statex \textbf{Output:} published unlearned parameters $\tilde{\theta}_i$.
\Procedure{PGDUnlearn}{$\tilde{\theta}_{i-1}$, $D_{i-1}$, $\mathbf{z}_{i-1}$; $T_i$, $\sigma$}
\State initialise $\theta^{\prime}_{0}\leftarrow \tilde{\theta}_{i-1}$
\State $D_{i}\leftarrow D_{i - 1} \setminus \{\mathbf{z}_{i-1}\}$
\For{$t = 1; t\leq T_i; t++$}
    \State $\theta^{\prime}_{t}\leftarrow \operatorname{Proj}_{\mathrm{\Theta}}(\theta^{\prime}_{t-1} - \eta_{t}\nabla L(\theta^{\prime}_{t - 1}, D_i))$
\EndFor
\State $\hat{\theta}_{i}\leftarrow \theta^{\prime}_{T_{i}}$ \label{alg:d2d-bpgd:line:secret-param}
\State draw $Z \sim \mathcal{N}(0, \sigma^{2}I_{p})$
\State \Return $\tilde{\theta}_{i} = \hat{\theta}_{i} + Z$
\EndProcedure
\end{algorithmic}
\end{algorithm}

\paragraph{Perturbed distributed descent.}

\citet{Neel_2020} also introduce a second algorithm, perturbed distributed descent, following the same vein as the work of \citet{zhang2013communication} and \citet{Bourtoule_2021}, suitable for larger datasets but that requires modifying the training process.
The algorithm partitions the data into $K$ parts, trains separate models on each partition, averages the parameters across the trained models, and then perturbs the final parameters.
Upon each deletion request, the models affected may be retrained using the basic perturbed gradient descent algorithm described above.
This may offer some improvements over \citet{Bourtoule_2021} by removing the need for na\"ive retraining of the affected model.
The authors also introduce a form of reservoir sampling prior to the gradient descent updates to ensure that the data post-deletion is distributed as i.i.d. samples drawn from the current dataset with replacement.
This is required for the guarantees on the out-of-sample error rate, described in \citet{zhang2013communication}, to hold.
In addition, \citet{Neel_2020} extend their algorithm by training $C$ different copies in parallel and publishing the one that achieves the lowest loss, in order to achieve stronger guarantees on consistency of the unlearned model.
The pseudocode for these procedures can be found in \citet[Algorithms 5, 6, 7]{Neel_2020}.

\subsubsection{Performance.}

\citet{Neel_2020} provide theoretical performance of the algorithms and do not provide any empirical evaluations.
However, the theoretical guarantees performances and guarantees are explicit and demonstrate clearly the role of parameters in controlling the trade-offs between efficiency, consistency and certifiability.
The performance of the algorithms depends on the convexity, smoothness, and Lipschitz property of the loss function \citep[see][Lemma 2.12]{Neel_2020} as well the required unlearning guarantees.

\paragraph{Efficiency.}

The authors measure efficiency in terms of the gradient descent iterations required for the $i$th deletion request relative to the run-time required for the first request, $\mathcal{I}$. 
For example, the imperfect strong perturbed gradient descent algorithm requires $\log(\frac{\epsilon n}{\sqrt{p}})$ fewer iterations than na\"ive retraining and the perfect version requires $\mathcal{I} + \log(\frac{\epsilon n}{\sqrt{p}}) - \log(i) \mathcal{I}$ fewer iterations, where $n$ is the dataset size and $p$ is the dimensionality.

\paragraph{Consistency.}

The $(\alpha, \beta)$-accuracy (see \cref{def:abacc}) guarantees of these unlearning methods depend on whether they are required to be strong or weak. 
The $(\alpha, \beta)$-accuracy of the distributed methods are analysed in depth by \citep{zhang2013communication}. 
For the perturbed gradient descent method, the $(\alpha, \beta)$-accuracy ranges from $\frac{pe^{-\mathcal{I}}}{\epsilon^{2}n^{2}}$ for the unregularized version to $(\frac{\sqrt{p}}{\epsilon n \sqrt{\mathcal{I}}})^{2/5}$ for the strong-regularized version. 

\paragraph{Certifiability.}

The authors prove the certifiability of the perfect and imperfect versions of their algorithms.
Given $\epsilon$ and $\delta$, \citet[Theorems 3.1, 3.2, 3.4, 3.5]{Neel_2020} define an explicit choice for the noise parameter $\sigma$, which depends on values for the smoothness, convexity, and Lipschitz constant of the loss function as well as $\epsilon$ and $\delta$.
They prove that this choice of $\sigma$ guarantees $(\epsilon, \delta)$-certifiability.

% -------------------------------------------------------------------------------------------------------------------
\subsection{DeepObliviate}

\label{sec:deepobliviate}

DeepObliviate is an approximate unlearning algorithm developed by \citet{he2021deepobliviate} that applies broadly to deep neural networks and follows a similar procedure to the slicing component of SISA, \Cref{sec:sisa}. The method further improves on slicing by using the so-called \textit{temporal residual memory} to identify which intermediate models need to be retrained, adding approximation into the process. 

\subsubsection{Methodology.}
DeepObliviate modifies conventional model training by dividing $D$ into $B$ disjoint subsets $\{D_1, \dots, D_B\}$, so that $D = \bigcup_i D_i$ and $\bigcap_i D_i = \varnothing$, and performs multi-epoch model training on each data block successively. Data blocks are uniform in two respects: (\textit{i}) each block is of size $|D|/B$ or $|D|/(B+1)$, and (\textit{ii}) the number of data points with the same label is uniform in each block. Model parameters are saved after each training block and are used as the initial parameters for the next training block, giving $B$ intermediate models $\{h_1, \dots, h_B\}$, with their associated parameters $\{\theta_1, \dots, \theta_B\}$ being saved during training. All parameter vectors are assumed to be in a fixed topological order so that weights corresponding to neural network edges going from the $i$th layer have smaller vector indices than those going from the $j$th layer, when $i<j$.

\begin{figure}
	\centering
	\includegraphics[scale=0.3]{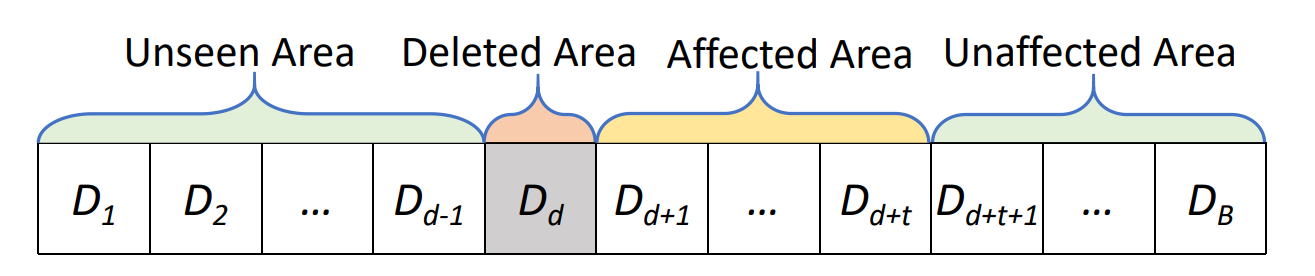}
	\caption{Unlearning in DeepObliviate where deleted data resides in block $D_d$ \citep{he2021deepobliviate}.\label{fig:DeepObliviate}}
\end{figure}

DeepObliviate identifies affected intermediate models that need to be retrained via the \textit{temporal residual memory}, defined below. Given a data point to unlearn, $\mathbf{z}\in D_d$, belonging in the $d$th data block ($1\leq d\leq B$), \citet{he2021deepobliviate} empirically observe that the temporal residual memory of $\mathbf{z}$ on $h_d$ is relatively large, but decays exponentially on successive models $h_{d+1}, \dots, h_B$. As such, the blocks of data can be divided into four distinct groups, see Fig. \ref{fig:DeepObliviate}, as follows: (\textit{i}) \textit{unseen area}, where the data point was not included in model training; (\textit{ii}) \textit{deleted area}, which includes the data to be deleted, $\mathbf{z}$, and where model retraining needs to start from; (\textit{iii}) \textit{affected area}, covering the models with prominent residual memory which also need to be fully retrained; and finally (\textit{iv}) \textit{unaffected area}, where the residual memory decay stabilises to the point that models do not need to be retrained and the remaining models can be easily approximated using the original intermediate models. This approach gives a probabilistic guarantee of speed-up over the na\"ive approach, since it is unlikely that retraining occurs on all blocks. 

\noindent We let $\textsc{Train}(\{D_1, \dots, D_B\}\mid h) := (\textsc{Train}(D_B\mid\cdot)\circ\cdots\circ\textsc{Train}(D_1\mid\cdot))(h)$ denote recursively training the initialised model $h$ on the datasets $D_1, \dots, D_B$. The unlearning procedure can then be summarised as:
\begin{equation}
h^{\mathcal{U}} = \textsc{Train}(\{D_d', D_{d+1}, \dots, D_{d+t}\}|h_{d-1})\oplus(h_B\ominus h_{d+t}),
\end{equation}
where $h^{\mathcal{U}}$ is the unlearned model, $D_d' = D_d\setminus\{\mathbf{z}\}$ excludes the unlearning data from $D_d$, $\oplus$ and $\ominus$ implement vector addition and subtraction on the corresponding model parameters, and $t$ is the number of data blocks after which unlearning influence can be ignored and so retraining can stop.

The value of $t$ is determined by the temporal residual memory. \citet{he2021deepobliviate} define the temporal residual memory as the $\ell^1$ distance (or Manhattan distance) between the influence of the deleted data $\mathbf{z}$ on successive models when $\mathbf{z}$ is included in training and when it is not. Formally, the temporal residual memory $\mathrm{\Delta}_t$ at step $t$ is:
\begin{align}
\mathrm{\Delta}_t := || I( D_{d+t} | h_{d+t-1} ) &- I( D_{d+t} | h^{\mathcal{U}}_{d+t-1} ) ||_1,\\
I( D_i | h_{i-1} ) :&= h_i \ominus h_{i-1},
\end{align}
where $0\leq t\leq B-d$, $I$ denotes the \emph{temporal influence} of block $D_i$ on $h_{i-1}$, and $\|\cdot\|_1$ is the $\ell^1$ norm (i.e., $\|(x_1, \dots, x_N)\|_1 = |x_1| + \cdots + |x_N|$) . Once $\mathrm{\Delta}$ is small enough and stable, retraining can stop, i.e., the influence of deleted data can be ignored for downstream models.

\citet{he2021deepobliviate} use \textit{detrended fluctuation analysis} or DFA \citep{Peng1994} to eliminate noise in $\mathrm{\Delta}$ and systematically determine whether $\mathrm{\Delta}$ has stabilised. DFA is used to determine the statistical self-affinity of a time-series signal by fitting $\mathrm{\Delta}_t$ with a decaying power-law function, whose derivative is easily-computable and can be used to determine stationarity.

\begin{algorithm}
	\caption{Unlearning with DeepObliviate, \citep{he2021deepobliviate}.} 
	\label{alg:deepobliviate}
	\begin{algorithmic}[1]
		\Statex \textbf{Input:} parameters for intermediary trained models $\boldsymbol{\theta} = \{\theta_1, \dots, \theta_B\}$, training data $D = D_1\cup \cdots 
\cup D_B$, data point to be removed $\mathbf{z}\in D_d$, stationarity hyperparameter $\varepsilon$.
		\Statex \textbf{Output:} unlearned model $h^{\mathcal{U}}$.
		\Procedure{DeepObliviateUnlearn}{$\boldsymbol{\theta}$, $D$, $\mathbf{z}$; $\varepsilon$}
		\State initialise $h^{\mathcal{U}} \leftarrow h_{d-1}$
		\State initialise $\theta_{d-1}^{\mathcal{U}} \leftarrow \theta_{d-1}$
		\State $D_d\leftarrow D_d\setminus\{\mathbf{z}\}$
		\For {$t = 0; t \leq B-d; t++$} 
			\State $h^{\mathcal{U}} \leftarrow\textsc{Train}(D_{d+t}\mid h^{\mathcal{U}})$
			\State $\theta_{d+t}^{\mathcal{U}} \leftarrow h^{\mathcal{U}}$ get parameters from $h^{\mathcal{U}}$
			\State compute temporal influence $V_{d+t}\leftarrow\theta_{d+t} - \theta_{d+t-1}$ of $D_{d+t}$ on $\theta_{d+t-1}$
			\State compute temporal influence $V_{d+t}^{\mathcal{U}}\leftarrow\theta_{d+t}^{\mathcal{U}} - \theta_{d+t-1}^{\mathcal{U}}$ of $D_{d+t}$ on $\theta_{d+t-1}^{\mathcal{U}}$
			\State compute temporal residual memory $\mathrm{\Delta}_{d+t}\leftarrow|| V_{d+t} - V_{d+t}^{\mathcal{U}} ||_{1}$
			\State compute power-law exponent using DFA $\alpha\leftarrow \textsc{DFA}( \{\mathrm{\Delta}_{d}, \mathrm{\Delta}_{d+1}, \dots, \mathrm{\Delta}_{d+t} \} )$
			\State $Y(x) := ax^{-\alpha} + b$
			\State $f(x) := \partial Y(x) / \partial x = a\cdot(-\alpha)\cdot x ^{-\alpha-1}$
			\State $a, b\leftarrow\argmin_{a, b}(Y(x) - \mathrm{\Delta}_{x})$ using least  squares, $x\in\{d, \dots, d+t\}$
			\State $g\leftarrow f(d+t)$
			\If {$|g| < \varepsilon$}
				\State\textbf{break}
			\EndIf
		\EndFor
		\State $h^{\mathcal{U}} \leftarrow h^{\mathcal{U}} \oplus (h_{B} \ominus h_{d+t})$
		\State $\{\theta_{d}, \dots, \theta_{d+t}\}\leftarrow\{\theta_{d}^{\mathcal{U}}, \dots, \theta_{d+t}^{\mathcal{U}}\}$
		\State \textbf{Return} $h^{\mathcal{U}}$
\EndProcedure
	\end{algorithmic} 
\end{algorithm}	

\subsubsection{Performance.}
\citet{he2021deepobliviate} evaluate the unlearning method on five public datasets: \textsc{mnist}, \textsc{cifar}-10, \textsc{svhn}, Purchase and ImageNet. Their evaluation considers ranges of volumes of deletion data and of block position for the unlearned data. We summarise performances for the deletion of a single data point; all performances degrade for larger volumes of deletion, though consistency is more robust to increases in deletion volume than efficiency, effectiveness and certifiability. Full results can be found in \citet[Tables II-VI]{he2021deepobliviate}. Results are also benchmarked in (\emph{ibid.}, Table VII) against the SISA method of \Cref{sec:sisa} by selecting the number of shards $S$ and slices $R$ so that $B = SR$.

\paragraph{Efficiency.} DeepObliviate is shown to achieve $\speed$ of 75.0$\times$, 66.7$\times$, 33.3$\times$, 29.4$\times$, 13.7$\times$ on \textsc{svhn}, \textsc{mnist}, \textsc{cifar}-10, Purchase, and ImageNet datasets, respectively. It is shown in \citet[Table VII]{he2021deepobliviate} that DeepObliviate has significant efficiency improvements over the SISA method for all datasets.

The hyperparameter $\varepsilon$ determines the minimum stationarity value of the residual memory decay, under which retraining procedure stops. Therefore it is an efficiency parameter, with larger values increasing the efficiency of the DeepObliviate. This can be observed in the $\varepsilon$ values shown in \emph{ibid.} (Tables II-V).

\paragraph{Effectiveness.} $\eff$ across datasets is $<1\%$. DeepObliviate achieves better values for $\eff$ than SISA for all datasets. Increasing $\varepsilon$ will degrade $\eff$.

\paragraph{Consistency.} Consistency is measured using $\outputcon$, which measures the proportion of predictions that the unlearned model and na\"ive retrained model agree on. $\outputcon$ ranges from $\sim$96\% to $>99\%$.

\paragraph{Certifiability}
Certifiability of DeepObliviate is measured using the backdoor verification method described in \Cref{sec:certifiability} and \citet{sommer2020probabilistic}. As noted in \Cref{sec:certifiability}, this is comparable to the use of $\certdis$ but ensures that na\"ive retraining has poor predictive performance on the deleted data, making $\certdis$ more meaningful. To compute $\certdis$, 500-1,000 data points are deleted and accuracies of predicting on the augmented deleted data are reported for both DeepObliviate and na\"ive retraining. For $\varepsilon = 0.05$, this gives a $\certdis$ range of $[0.9\%, 10\%]$. Certifiability decreases for larger $\varepsilon$, demonstrating the efficiency-certifiability trade-off, e.g., $\certdis\in[2.7\%, 12.5\%]$ for $\varepsilon = 0.1$.

\section{Discussion}
\label{sec:discussion}

In this section, we discuss and compare the seven methods described in the previous sections. In addition, we discuss some aspects of moving the field of machine unlearning towards practice, framed around approaches for empirical method selection and method monitoring. Table \ref{table:datasets} gives statistics on open-source datasets that have been used in the literature to obtain experimental results for the seven methods we have reported on. Many of these datasets are commonly-used benchmarking datasets across the machine learning literature; further descriptions of the datasets can be found in Appendix \ref{appendix:datasets}.

\begin{table}
\resizebox{0.92\textwidth}{!}{\begin{minipage}{\textwidth}
\centering
\caption{\label{table:datasets} Summary of the datasets used in the experiments of the papers discussed. \textit{Dimensionality} refers to the number of prediction features. When datasets have separate and designated training and testing splits, \textit{Size} refers to the number of samples in the training dataset, else it gives the number of samples in the whole dataset. \textit{Class balance} is given only for datasets with binary label and reports the percentage of positive labels amongst the number of samples in the \textit{Size} column. Descriptions of the datasets can be found in Appendix \ref{appendix:datasets}.}
\begin{tabular}{|c|c|c|c|c|c|}
	\hline
	\multirow{2}{*}{\textbf{Dataset}} & \multirow{2}{*}{\textbf{Dimensionality}} & \multirow{2}{*}{\textbf{Size}} & \multirow{2}{*}{\textbf{Classes}} & \multirow{2}{*}{\textbf{Class balance}} & \multirow{2}{*}{\textbf{Appears in}}\\
	& & & & & \\ 
	\hline
	\multirow{3}{*}{\textsl{\textsc{mnist}}} & \multirow{3}{*}{784} & \multirow{3}{*}{60,000}& \multirow{3}{*}{10} & \multirow{3}{*}{N/A} & \citep{Mahadevan_2021},  \\
	& & & & & \citep{guo2020certified}, \citep{Wu_2020},  \\
	& & & & & \citep{Golatkar_2019}, \citep{he2021deepobliviate} \\
	\hline
	\multirow{2}{*}{Covtype} & \multirow{2}{*}{54} & \multirow{2}{*}{581,012} & \multirow{2}{*}{7} & \multirow{2}{*}{N/A} & \citep{Mahadevan_2021}, \\
	& & & & & \citep{Wu_2020} \\
	\hline
	\multirow{2}{*}{\textsc{higgs}} & \multirow{2}{*}{28} & \multirow{2}{*}{11,000,000} & \multirow{2}{*}{2} & \multirow{2}{*}{$53.00\%$} & \citep{Mahadevan_2021}, \\
	& & & & & \citep{Brophy_2021}, \citep{Wu_2020} \\
	\hline
	Epsilon & 2,000 & 400,000 & 2 & $50.00\%$ & \citep{Mahadevan_2021} \\
	\hline
	\textsc{cifar-2} & 3,072 & 12,000& 2 & 50.00$\%$ & \citep{Mahadevan_2021}\\
	\hline
	\textsc{cifar-10} & 3,072 & 60,000& 10 & N/A & \citep{Golatkar_2019}, \citep{he2021deepobliviate} \\
	\hline
	\textsc{cifar-100} & 3,072 & 60,000 & 100 & N/A & \citep{Bourtoule_2021} \\
	\hline
	\textsc{lsun}  & 65,536 & 1,000,000 & 10 & N/A & \citep{guo2020certified} \\
	\hline
	\textsc{sst} & 215,154 & 11,855 & 5 & N/A & \citep{guo2020certified} \\
	\hline
	\textsc{svhn} & 3,072 & 73,257  & 10 & N/A & \citep{guo2020certified}, \citep{he2021deepobliviate} \\
	\hline
	\textsc{rcv1}& 47,236 & 20,242 & 2 & $52.00\%$& \citep{Wu_2020}\\
	\hline
	Purchase & 600 & 250,000 & 2 &N/A & \citep{Bourtoule_2021} \\
	\hline
	ImageNet & 150,528 & 1,281,167  & 1000 & N/A & \citep{Bourtoule_2021}, \citep{he2021deepobliviate}\\
	\hline
	Mini-ImageNet  & 150,528 & 128,545 & 100 & N/A & \citep{Bourtoule_2021}\\
	\hline
	Surgical & 25 & 14,635 & 2 & $25.24\%$ & \citep{Brophy_2021} \\
	\hline
	Vaccine & 36 & 26,707 & 2 & $46.40\%$& \citep{Brophy_2021} \\
	\hline
	Adult & 14 & 48,842 & 2 & 23.92$\%$ & \citep{Brophy_2021}\\
	\hline
	Bank Mktg. & 17 & 45,211 & 2 & $11.30\%$ & \citep{Brophy_2021} \\
	\hline
	Diabetes  & 20 & 101,766 & 2 & $46.10\%$ & \citep{Brophy_2021}\\
	\hline
	No Show & 14 & 110,527 & 2 & $20.19\%$& \citep{Brophy_2021} \\
	\hline 
	Olympics & 15 & 206,165 & 2 & $14.60\%$ & \citep{Brophy_2021}\\
	\hline
	Census & 40 & 299,285 & 2 & $6.20\%$& \citep{Brophy_2021} \\
	\hline
	Credit Card  & 30 & 284,807 & 2 & $0.17\%$ & \citep{Brophy_2021}\\
	\hline
	\textsc{ctr} & 39 & 1,000,000 & 2 & 2.91$\%$ & \citep{Brophy_2021}\\
	\hline
	Twitter & 15 & 1,000,000 & 2 & $16.93\%$ & \citep{Brophy_2021}\\
	\hline
	Lacuna-10 & 1,024 & $\geq 5,000$ & 10 & N/A & \citep{Golatkar_2019}\\
	\hline
	Lacuna-100 & 1,024 & $\geq 50,000$ & 100 & N/A & \citep{Golatkar_2019} \\	
	\hline

\end{tabular}
      \end{minipage}}
\end{table}

Additionally, we provide two tables (Table \ref{table:summary} and Table \ref{table:performance}) that summarise the algorithms that we have described in this paper. These tables are summaries and should be used as a guide to the performances of individual algorithms, as reported by the relevant literature. In particular, they should not generally be used for comparisons between algorithms. Table \ref{table:summary} provides a summary of the methods considered in this paper, which categorises the various methods according to their applicability and certificate of unlearning, including some empirical results of certifiability. Table \ref{table:performance} summarises the empirical and theoretical results demonstrated in the literature. We can observe the efficiency-effectiveness trade-off in Table \ref{table:performance} for methods with multiple efficiency parameter values. For example, the bottom row of DaRE, corresponding to highest-tested $d_{\rmax}$, sees lower effectiveness values but higher efficiency than the top row, corresponding to lowest-tested $d_{\rmax}$. Likewise for DeepObliviate, Fisher, and Influence. We observe an efficiency-consistency trade-off for DeepObliviate, and an efficiency-certifiability trade-off in the Certificate of Unlearning column for DeepObliviate, Fisher, and Influence.

{\footnotesize
\begin{table}
\begin{center}
\begin{threeparttable}
\caption{\label{table:summary} Summary table of unlearning algorithms considered in this paper. In the table, SC = strongly convex loss function, BG = bounded gradients for loss function \citep[Assumption 4]{Wu_2020}, SI = strong independence \citep[Assumption 5]{Wu_2020}, LQ = locally quadratic. This table is a summary only and empirical results contained in the table should not be used for comparison due to the use of different datasets, experimental design, and machine specifications.}
\begin{tabularx}{\textwidth}{|c|c|c|Y|Y|}
	\hline
	\multicolumn{5}{|c|}{\textbf{Unlearning Type}} \\
	\hline
	\multirow{2}{*}{\textbf{Method}} & \multirow{2}{*}{\textbf{Applicability}} & \multirow{2}{*}{\textbf{Properties}} & \multicolumn{2}{c|}{\textbf{Certificate}}  \\
	& &  & \multicolumn{2}{c|}{\textbf{of Unlearning}} \\
	\hline 
	\multirow{2}{*}{SISA} & Incrementally trained & Exact & \multicolumn{2}{c|}{\multirow{2}{*}{Exact}}  \\
	 & models$^\textbf{a}$ & Weak$^\textbf{b}$ & \multicolumn{2}{c|}{} \\
	\hline
	\multirow{2}{*}{DaRE} & DaRE trees & \multirow{2}{*}{Exact} & \multicolumn{2}{c|}{\multirow{2}{*}{Exact}}  \\
	 & DaRE RF & & \multicolumn{2}{c|}{}  \\
	\hline
	\multirow{2}{*}{Fisher} & LQ loss & \multirow{2}{*}{Approximate} & \multirow{2}{*}{3.3, 33.4 kNATs$^{\textbf{c}}$} & $0.0\%^{\textbf{d}}$ ($m' = \lfloor m/8\rfloor$) \\
	 & function & & & $0.26\%$ ($m' = m$)\\ 
	\hline
	\multirow{2}{*}{Influence} & SC, BG, & Approximate & \multirow{2}{*}{($\epsilon, \delta)$-certified} & $0.1\%^{\textbf{d}}$ ($m' = \lfloor m/8 \rfloor$) \\
	& Lipschitz Hessian & Weak$^\textbf{e}$ & & $1.1\%$ ($m' = m$) \\
	\hline 
	\multirow{2}{*}{DeltaGrad} & SC, smooth loss, BG, & Approximate & \multicolumn{2}{c|}{\multirow{2}{*}{N/A$^\textbf{g}$}}  \\
	 & Lipschitz Hessian, SI & Strong$^\textbf{f}$ & \multicolumn{2}{c|}{} \\
	\hline
	Descent-to- & SC, smooth & Approximate & \multicolumn{2}{c|}{\multirow{2}{*}{$(\epsilon, \delta)$-certified}}  \\
	Delete & (Bounded, Lipschitz Hessian)$^\textbf{h}$ & Strong$^{\textbf{i}}$& \multicolumn{2}{c|}{} \\ 
	\hline
	\multirow{2}{*}{DeepObliviate} & Any deep-learning & Approximate & \multicolumn{2}{c|}{$6.39\%$ ($\varepsilon = 0.05$)}\\ 
	& model & Strong & \multicolumn{2}{c|}{$8.28\%$ ($\varepsilon = 0.1$)$^\textbf{j}$}\\
	\hline 
\end{tabularx}
\begin{tablenotes}
	\item \textbf{a.} Sharding-only SISA is applicable to any machine learning model, including decision-tree based models; the full SISA algorithm is applicable to any model that has been trained incrementally, for example via gradient descent.
	\item \textbf{b.} See Equations (6) and (8), of \citet[Section V.C.]{Bourtoule_2021}.
	\item \textbf{c.} Certifiability of Fisher in \citet{Golatkar_2019} is given by an upper bound on the information retained. Empirical results are given only for two datasets, Lacuna-10 and CIFAR-10, with the model retaining at most 3.3 kNATs of information 33.4 kNATs of information, respectively, about the deleted data. 
	\item\textbf{d.} Certifiability of Fisher and Influence in \citet{Mahadevan_2021} is given by $\certdis$. We report the geometric mean across datasets for two extremes of the efficiency parameter $m'$, fixing $\sigma = 1$ and for the smallest deletion volume $m$ considered.
	\item \textbf{e.} The convergence error grows linearly with number of points removed, see \citep[p.4]{guo2020certified}.
	\item \textbf{f.} When deleting a single point (i.e., $r = 1$) the convergence rate is $o(1/n)$ regardless of position in sequence \citep[see][Theorem 1]{Wu_2020}. This is expected since DeltaGrad computes gradient descent from scratch each time.
	\item \textbf{g.} $\certdis$ is used in \citet{Mahadevan_2021}, but certifiability is not measured in the original paper \citep{Wu_2020}.
	\item \textbf{h.} Strongly convex can be relaxed to convex by using regularised perturbed gradient descent  \citep[Theorems 3.4, 3.5]{Neel_2020}. Bracketed assumptions are additional assumptions required for distributed perturbed gradient descent.
	\item \textbf{i.} All descent-to-delete algorithms are strong apart from that of \citet[Theorem 3.5]{Neel_2020}, which is designed to be weak in order to allow better consistency.	
	\item \textbf{j.} Certifiability values for DeepObliviate are taken from \citet[Table VI]{he2021deepobliviate} which applies the backdoor verification experiment. We show the two different values of the efficiency parameter $\varepsilon$, and assume the deleted data is in the first block (top row of each dataset in Table VI, \emph{ibid.}). Values are then passed through $\certdis$, and we report the geometric mean across all tested datasets.	
\end{tablenotes}
\end{threeparttable}
\end{center}
\end{table}
\begin{table}
\begin{center}
\begin{threeparttable}
\caption{\label{table:performance} Summary of efficiency, effectiveness, and consistency performances of the algorithms considered as reported by the literature. Note that this table is a summary only and should not be used for comparison, since the results are obtained using different datasets, experimental design and machine specifications.}
\begin{tabularx}{\textwidth}{|Y|Y|c|c|c|c|c|c|c|}
	\hline
	\multicolumn{9}{|c|}{\textbf{Performance}} \\
	\hline
	\multicolumn{2}{|c|}{\multirow{2}{*}{\textbf{Method}}} & \multicolumn{3}{c|}{\textbf{Efficiency}}
& \multirow{2}{*}{\textbf{Effectiveness}} & \multicolumn{3}{c|}{\textbf{Consistency}} \\ \cline{3-5} \cline{7-9}
	\multicolumn{2}{|c|}{} & \textbf{min} & \textbf{g. mean} & \textbf{max} &  & \textbf{min} & \textbf{g. mean} & \textbf{max}  \\
	\hline
	\multirow{2}{*}{SISA$^\textbf{a}$} & \citet{Bourtoule_2021} & 1.36$\times$ & 2.49$\times$ & 4.63$\times$ & $< 2\%$ ($18.76\%$) & \multicolumn{3}{c|}{\multirow{2}{*}{Exact}}\\  \cline{2-6}
	 & \citet{he2021deepobliviate} & 14.0$\times$ & 39.6$\times$ & 75.0$\times$ & $<5\%$ ($15.1\%$) & \multicolumn{3}{c|}{} \\
	\hline
	\multirow{2}{*}{DaRE$^\textbf{b}$} & min $d_{\rmax}$ & 10$\times$ & 366$\times$ & 9735$\times$ & $\leq 0.5\%$ & \multicolumn{3}{c|}{\multirow{2}{*}{Exact}} \\ \cline{2-6}
	& max $d_{\rmax}$ & 145$\times$ & 1272$\times$ & 35856$\times$ & $\leq 2.5\%$ & \multicolumn{3}{c|}{} \\
	\hline
	\multirow{2}{*}{Fisher$^{\textbf{c}}$} & $m' = \lfloor m/8\rfloor$ & 1.0$\times$ & 3.3$\times$ & 36.8$\times$ & $\approx 0.0\%$ & \multicolumn{3}{c|}{\multirow{2}{*}{N/A}} \\ \cline{2-6}
	& $m' = m$ & 1.5$\times$ & 13.0$\times$ & 282.4$\times$ & $<0.2\%$ & \multicolumn{3}{c|}{}\\ 
	\hline
	\multirow{2}{*}{Influence$^{\textbf{c}}$} & $m' = \lfloor m/8\rfloor$ & 0.3$\times$ & 5.9$\times$ & 75.7$\times$ &$< 0.55\%$ &\multicolumn{3}{c|}{\multirow{2}{*}{N/A}}  \\ \cline{2-6}
	& $m' = m$ & 2.5$\times$ & 38.2$\times$ & 215.1$\times$ & $<0.57\%$ & \multicolumn{3}{c|}{}\\
	\hline
	\multicolumn{2}{|c|}{\multirow{2}{*}{DeltaGrad$^{\textbf{d}}$}} & \multirow{2}{*}{1.6$\times$} & \multirow{2}{*}{2.7$\times$} & \multirow{2}{*}{6.5$\times$} & \multirow{2}{*}{$\approx 0.0\%$} & \multirow{2}{*}{$1.7\times 10^{-6}$} & \multirow{2}{*}{$1.1\times 10^{-5}$} & \multirow{2}{*}{$1.4\times 10^{-4}$} \\
	\multicolumn{2}{|c|}{} & & & & & & &  \\
	\hline
 	& PGD & \multicolumn{3}{c|}{$1 + \mathcal{I}^{-1}\log(\epsilon n/\sqrt{p})$} & \multirow{5}{*}{N/A}&\multicolumn{3}{c|}{$ p/(e^{\mathcal{I}}\epsilon^2n^2)$} \\ \cline{2-5}\cline{7-9}
	Descent-to- & sRPGD & \multicolumn{3}{c|}{$\mathcal{I}^{-\frac{3}{5}}(\epsilon n/\sqrt{p})^{\frac{2}{5}}$} & & \multicolumn{3}{c|}{$\left(\frac{\sqrt{p}}{\epsilon n\mathcal{I}}\right)^{\frac{2}{5}}$} \\ \cline{2-5}\cline{7-9}
	Delete$^{\textbf{e}}$ & wRPGD  & \multicolumn{3}{c|}{$\mathcal{I}^{-\frac{3}{4}}\sqrt{\epsilon n/\sqrt{p}}$} & & \multicolumn{3}{c|}{$\sqrt{\frac{\sqrt{p}}{\epsilon p\sqrt{\mathcal{I}}}}$} \\ \cline{2-5}\cline{7-9}
	\multirow{2}{*}{}& \multirow{2}{*}{DPGD} & \multicolumn{3}{c|}{$\min\{\mathcal{I}^{-1}\log n, $}  & \multirow{2}{*}{}& \multicolumn{3}{c|}{\multirow{2}{*}{$\frac{2\exp\left(-\mathcal{I}n^{\frac{4-3\xi}{2}}\right)}{\epsilon^2 n^2} + \frac{1}{n^{\xi}}$}}\\
	& & \multicolumn{3}{c|}{$n^{\frac{4-3\xi}{2}} + \mathcal{I}^{-1}\log(\epsilon n/\sqrt{p})\}$} & & \multicolumn{3}{c|}{} \\
	\hline
	\multirow{2}{*}{DeepObliviate$^\textbf{f}$} & min $\varepsilon$ & 6.17$\times$ & 13.97$\times$ & 45.45$\times$ & $<0.18\%$ & 97.25 & 98.82 & 99.95\\ \cline{2-9}
	& $\varepsilon=0.1$ & 9.26$\times$ & 22.81$\times$ & 66.67$\times$ & $<0.35\%$ & 95.78 & 97.61 &  99.85 \\
	\hline
\end{tabularx}
\begin{tablenotes}
	\item \textbf{a.} The top row gives results from the original paper \citep{Bourtoule_2021}. Results are reported for multiple deletions (8, 18, or 39) depending on the dataset. Effectiveness is $<2\%$ for simpler learning tasks and $18.76\%$ for ImageNet. The second row gives results from \citet{he2021deepobliviate}, which is for single deletions, and give $15.1\%$ effectiveness reported for ImageNet.
	\item \textbf{b.} Actual value of minimum and maximum $d_{rmax}$ depends on the dataset, and determined with respect to CV error tolerances of $0.1\%$ and $1.0\%$, respectively, in cross-validation, see Section \ref{sec:dare} and \citet[Table 6]{Brophy_2021}. 
	\item \textbf{c.} Results are taken from \citet{Mahadevan_2021} for a fixed noise parameter $\sigma = 1$ and for the smallest volume of data deleted, which volume depends on the specific dataset. The parameter $m'$ is the mini-batch size used in the update, and $m$ is the volume of data deleted.
	\item \textbf{d.} Consistency of DeltaGrad is measured by $\paramcon$; results are from \citet{Wu_2020}.
	\item \textbf{e.} PGD = perturbed gradient descent; sRPGD/wRPGD = strong/weak variant of regularized perturbed gradient descent; DPGD = distributed perturbed gradient descent. Results are extracted from \citet[Table 1]{Neel_2020}. Efficiency is given by reporting the number of iterations for na\"{i}ve retraining divided by the number of iterations for the first deletion $\mathcal{I}$. $\epsilon$ is the value in the $(\epsilon, \delta$)-certified guarantee of the method. $\xi\in[1, 4/3]$ is a parameter in distributed descent. Consistency is given by $(\alpha, \beta)$-accuracy (Definition \ref{def:abacc}). 
	\item \textbf{f.} In DeepObliviate, we take values from \citet[Tables I, II, III, V]{he2021deepobliviate}, corresponding to four experiments that use multiple $\varepsilon$ values (Tables IV and VII use only a single $\varepsilon$ value). Within each experiment, there are different choices of the number of data unlearned, the position of data to be unlearned and the choice of hyperparameter $\varepsilon$. We fix the smallest volume of data unlearned, and earliest block position to give worst-case efficiency results. We fix $\varepsilon$ at two extremes, and report for each; minimum $\varepsilon$ depends on the dataset (either 0.04 or 0.05) and maximum $\varepsilon$ is always 0.1. Consistency is measured by $\outputcon$.
\end{tablenotes}
\end{threeparttable}
\end{center}
\end{table}
}

\subsection{Exact Unlearning Algorithms} In Section \ref{sec:exact_methods}, we described SISA, whose full method applies to any incrementally-trained machine learning model, and DaRE forests, which applies to decision trees and random forests. Both SISA and DaRE forests each come with two efficiency parameters in order to control the unlearning efficiency, usually at the cost of effectiveness. However, unlike approximate unlearning methods, such parameters are also involved in the initial training procedure, which means that they can be tuned during training to maximise efficiency with respect to a satisficing bound on effectiveness.

\citet{Brophy_2021} obtained strong empirical results for DaRE forests, reporting an average value of $\speed$ of 2-3 orders of magnitude across the 14 tested datasets, up to a maximum of 5 orders of magnitude, when removing single data points. Results of a similar magnitude to DaRE are reported for another decision-tree based unlearning method in \citet{Schelter_2021}, which may suggest that decision tree models are more amenable to efficient unlearning. SISA achieves a maximum $\speed$ of 4.63$\times$ in the original paper \citep{Bourtoule_2021}, when unlearning 8 data points from the Purchase dataset. \citet{he2021deepobliviate} obtained $\speed$ of up to 73$\times$ for single removals with SISA. 

Exact methods have the benefit of a high degree of consistency and certifiability, as given by their proven equivalence with the na\"ive removal mechanism, however they may share with the corresponding na\"ive removal mechanisms a vulnerability to membership inference attacks on the deleted data. SISA and DaRE therefore require a level of trust in the security of the pre- and post-unlearning machine learning models. This could be alleviated by the addition of noise into the training procedure, like many approximate methods do, however this will likely degrade efficiency. While SISA is broadly applicable, it struggles to provide a significant efficiency or decent effectiveness for more complex learning tasks, as seen with the $\speed$ of 1.36$\times$ and $\eff$ of 18.76\% on ImageNet. The efficiency of DaRE is more pronounced on balanced datasets, such as \textsc{higgs}, and reports lower efficiencies on datasets with extreme class imbalances, e.g., Credit Card. The DaRE methodology is not currently applicable to boosting ensembles, suggesting a natural next step for the development of this algorithm. Finally, both exact methods come with additional storage costs for storing statistics and parameters during the training and unlearning; these storage operations may also incur additional computational cost, however, this is not explicitly mentioned in either \citet{Bourtoule_2021} or \citet{Brophy_2021}.

\subsection{Approximate Unlearning Algorithms} In Section \ref{sec:approx_methods}, five approximate algorithms were described. Three of these, Fisher, Influence, and Descent-to-Delete, apply parameter updates starting from the optimised parameters of the original machine learning model; Descent-to-Delete performs first-order updates, Fisher performs a second-order Newton update using the remaining data $D\setminus D_u$, and Influence performs a second-order Newton update using the deleted data $D_u$. DeltaGrad applies gradient descent updates starting from the same initialised parameters as the original training procedure, and approximates the gradient descent steps of the na\"ive unlearning mechanism by using cached historical training information. Finally, DeepObliviate takes the slicing procedure of the SISA unlearning algorithm and turns this into an approximate algorithm by stopping the retraining procedure at an appropriate point and approximating the tail end of the retraining by using cached parameters from the initial training procedure. The applicability of approximate unlearning algorithms is generally narrow. Fisher, Influence, DeltaGrad, and Descent-to-Delete are primarily applicable to models with strongly-convex loss functions, although extensions of Fisher, Influence, and DeltaGrad to non-convex models are discussed in their respective papers. DeepObliviate is notable for being widely applicable to any neural network.

Due to differing experimental design, direct comparisons between the five approximate algorithms are difficult. However, since \citet{Mahadevan_2021} benchmark Fisher, Influence, and DeltaGrad applied to a logistic regression model, we can make some inferences on their performances. In general, Influence observes better $\speed$ values than Fisher, which can be seen in the efficiency column of Table \ref{table:performance}. This was noted in \emph{ibid.}, with $\speed$ being most pronounced on high-dimensional datasets, \textsc{cifar2} and Epsilon, and less pronounced on the low-dimensional datasets, Covtype and \textsc{higgs}. However, the Influence method has weaker effectiveness than the Fisher method. Across most scenarios, the efficiency and effectiveness of DeltaGrad is outperformed by the Influence and Fisher algorithms. Note that the original DeltaGrad is supplemented by \cite{Mahadevan_2021} with the addition of noise, in order to guarantee a degree of certifiability, however this appears to degrade the efficiency results of DeltaGrad. An additional reason for DeltaGrad's poorer performance is due to the computational cost of SGD iterations in DeltaGrad being higher than inverting the Hessian in Fisher and Influence. DeepObliviate is directly benchmarked against the SISA method in \citet{he2021deepobliviate}, achieving better $\speed$ and $\eff$ than SISA in most cases. 

\cite{Mahadevan_2021} measure certifiability using the $\certdis$ measure of (\ref{certdis}) without the backdoor verification method. As discussed in Section \ref{sec:certifiability}, the backdoor verification method can make the $\certdis$ measure more meaningful, which is done by \cite{he2021deepobliviate}. Finally, Descent-to-Delete suffers from non-existent empirical evaluation. A thorough benchmarking of all seven algorithms would be a beneficial next step.

\subsection{Machine Unlearning from a Practitioner's Perspective}
\label{sec:selection}

The evaluation of a single machine unlearning algorithm is complex and multidimensional. On top of this, evaluation measures in the literature differ, making comparisons between established algorithms difficult. In this section, we explore a framework for machine unlearning algorithm selection.

\paragraph{Candidate selection.} 

The various columns of Tables \ref{table:summary} and \ref{table:performance} correspond to key factors to consider when selecting an unlearning algorithm. We can use Table \ref{table:summary} to select algorithms that are applicable with respect to initial constraints. The Applicability column indicates which machine learning models the algorithm can be applied to. For example, a tree-based model is required for DaRE forests.

The remaining columns of Table \ref{table:summary} are of a high priority if the unlearning method is being applied in order to comply with regulations concerning removal requests. This compliance depends on the particular external regulations, for example, in the EU, unlearn requests must comply with the user's Right to be Forgotten. Future clarification of these regulations may rule out approximate algorithms, or they may impose strict constraints on the certificate of unlearning. So it is a key responsibility of researchers and practitioners to define and select algorithms with a suitable certificate of unlearning. Another consideration here is in removal immediacy. If data points must be removed immediately on request, then the strong and weak classification becomes important, whereas if not, then methods that allow batch unlearning may be more appropriate.

\paragraph{Parameter trade-offs and method selection.}
Given a set of candidate methods that are all applicable to the use case, we now discuss how to select among them. Currently, direct comparisons between algorithms are difficult for two main reasons: (1) experiments in papers vary greatly and, with the exception of \citet{Mahadevan_2021}, there has been no extensive empirical benchmark comparison of algorithms; (2) evaluation of unlearning algorithms is multidimensional, with various trade-offs that must be navigated relevant to the use case. As a result, Table \ref{table:performance} should be viewed as a guide to each individual algorithm and care should be taken with direct comparisons. 

Because of the difficulties in direct comparison, it makes sense to perform an empirical algorithm selection procedure once at the start of the pipeline, as is the current standard for machine learning. We now discuss a proposal for this procedure. Due to the computational expense of evaluating the unlearning algorithms, this procedure is advisable only if multiple unlearning processes are expected as a result of sequential deletion requests. First, according to both user specification (for example, the type of machine learning model in production and intended application of unlearning) as well as external certifiability constraints (for example, regulations), a list of candidate unlearning algorithms should be produced as discussed above. Candidates are both applicable to the problem and sufficiently certifiable. Additionally, the user should set tolerances for empirical results for effectiveness and certifiability (along with consistency, if desired); these are the minimal acceptable levels one expects to see in results for each evaluation criterion (for example, an effectiveness tolerance of 2\% means that algorithms that achieve $<2\%$ for $\eff$ are desired).

Once candidates and preset tolerances are chosen, an empirical evaluation routine is performed for each of the candidate algorithms. To speed up the process, this process may be performed on a random subsample of the training data. If the candidate algorithm is exact then, as is done by \citet{Brophy_2021}, one may first tune the efficiency parameter so that efficiency is maximised with respect to the preset effectiveness tolerance. For example, with DaRE forests the parameter $d_{\rmax}$ may be incremented and the cross-validation performance of the resultant trained DaRE RF is measured. Once this performance degrades beyond the preset effectiveness tolerance, then the efficiency parameter is set. This is possible for exact methods since the efficiency parameter is a parameter in both the training and the removal mechanisms, and because there are strong guarantees for consistency and certifiability. 

Finally, a deletion distribution is chosen and then deletion points are drawn from the training data according to this distribution. If the candidate is exact then, using the tuned efficiency parameter, deletion points are unlearned and values for $\speed$ and $\eff$ are measured and reported. If the candidate is approximate, then an appropriate range of efficiency and certifiability parameters are chosen and, for each choice of parameters, values for $\speed$, $\eff$, $\outputcon$, and $\certdis$ are measured and reported. The user may then choose the algorithm that achieves the best $\speed$ whilst remaining within the preset tolerances for the other evaluation criteria.

\paragraph{Machine unlearning monitoring and auditing.} 

Effectiveness and certifiability (and sometimes consistency) are treated as satisficing metrics, so preset tolerances for each are implicit inputs in algorithm selection. At certain points it will be necessary to perform full na\"{i}ve retraining of the model. This is to ensure that, as the volume of deletions increases, effectiveness, consistency, and certifiability do not breach the preset tolerances. Effective unlearning monitoring is used to decide when it is appropriate to fully retrain. The issue, however, is that monitoring of all four areas of evaluation involves comparison with the na\"ive retrained model. Computing this baseline model after every deletion defeats the purpose of unlearning in the first place. Therefore it is necessary to introduce proxies to measure evaluation after deletions. Consistency is difficult to estimate, and is less of a concern if we have good estimates for the other three areas, so we do not focus on this here.

For simplicity, we assume that additions are not made to the training data during the deletion pipeline and that removals of only single data points are performed. Let $\mathcal{U}$ be a removal mechanism, $h_0 = h$ be the original machine learning model, and $D_0 = D$ be the training data for $h_0$. Suppose that we have a sequence of models $\{h_i\}_{i=0}^{m-1}$ and datasets $D_i$, where each $h_i = \mathcal{U}(h_{i-1}, D_{i-1}, \mathbf{z}_{i - 1})$ and $D_i = D_{i-1}\setminus\{\mathbf{z}_{i-1}\}$ are the results of unlearning a point $\mathbf{z}_i$ from the previous model. We assume that there is a subset of $k$ indices $\{i_j\}$ for which, after obtaining $h_{i_j}$, a na\"ive retraining has occurred, giving $k$ na\"ively retrained models $\{h_{i_j}^*\}_{j=1}^k$; the subsequent model $h_{i_j + 1} = \mathcal{U}(h_{i_j}^*, D_{i_j}, \mathbf{z}_{i_j})$ is obtained by unlearning on $h_{i_j}^*$. We explore how we might monitor the $m$th deletion.

For certifiability, we can overestimate $\certdis$ by the test error loss to the last retrained model as follows. Given a performance metric $\mathcal{M}$, let $\mathcal{M}_{\test, m}$ denote the performance of $h_m$ on a test set and let $\mathcal{M}_{\test, i_k}^*$ denote the test performance of $h_{i_k}^*$. We define
\[
	\widetilde{\certdis} := c_{i_k}\sape(\mathcal{M}_{\test, m}, \mathcal{M}_{\test, i_k}^*),
\]
where $c_{i_k}$ can be estimated empirically after each full retraining as shown in \citet[p. 13]{Mahadevan_2021}, and SAPE is as in (\ref{certdis}). This was shown, \emph{ibid.}, to give an overestimation of $\certdis$. 

In a similar way, we may overestimate $\eff$ by
\[
	\widetilde{\eff} := e_{i_k}|\mathcal{M}_{\test, m} - \mathcal{M}_{\test, i_k}^*|
\]
where $e_{i_k}$ is estimated empirically by replacing $\certdis$ in the calculation of $c_{i_k}$ with $\eff$. 

Efficiency is a key measure that should be monitored to ensure that the expected unlearning time is not consistently being exceeded, although it may not necessarily be used to decide when to retrain. We use the time taken to train the latest retrained model $h_{i_k}^*$ as a proxy for the time taken to fully retrain. Since we assume that there have been no additions to the dataset, we have $\len(D_{i_j}) > \len(D_m)$ so 
\[
	\widetilde{\speed} := \frac{\text{time taken to train $h_{i_k}^*$}}{\text{time taken to unlearn $\mathbf{z}_{m+1}$}}
\]
provides a conservative estimate of $\speed$.

After obtaining $h_{m}$, $\widetilde{\eff}$, and $\widetilde{\certdis}$ are calculated. If these values exceed a preset threshold for effectiveness and certifiability, then full retraining should occur to obtain $h_{i_{k+1}}^*$. Additionally, $\widetilde{\speed}$ should be calculated, and consistently poor $\widetilde{\speed}$ compared to those seen in the initial algorithm selection should be investigated. Also, unlearning algorithms with lower certifiability will not only be more likely to breach regulations, but they will need to be retrained more often as well, entailing higher cost in the long run. The estimates introduced here may also be applicable to the situation of unlearning algorithm selection described previously. 

Auditing may occur at the request of regulators. The precise formalisation of the certificate of unlearning in practice remains an open problem, but source code and enough information to recalculate the sequence of models $\{h_i\}_i$ above are likely to be requested. The regulator may then recreate the sequence of deletions and calculate certifiability according to their own measures, the failure of which will lead to fines or other regulatory actions. It is therefore important that effective monitoring and regular na\"ive retraining take place.

\section{Conclusion}
In this review paper, we give a broad introduction to and assessment of the field of machine unlearning. We provide a standardised unlearning and evaluation framework, along with the theory and implementations of seven state-of-the-art unlearning algorithms. Finally, in the Discussion section (Section \ref{sec:discussion}) we begin to address some of the theoretical and practical gaps identified in the field, however, more research is required to fully resolve these gaps. Whilst we are able to make some comparisons between the algorithms considered here, extensive empirical benchmarking between them is lacking in the literature, which could be a useful future contribution. In addition, there is a current lack of research into applying unlearning in practice and certifiability is a key concern in this regard. In order to verify that unlearning algorithms are removing sufficient information about the deleted data, a formalised certificate of unlearning that is applicable in practice, along with a robust monitoring pipeline, is a necessary piece of research, which is a sentiment also echoed elsewhere in the literature \citep{thudi2021necessity}. Finally, there is a general difficulty in developing unlearning algorithms that are applicable to modern deep neural networks, and more work in this area would be beneficial.

\subsubsection*{Acknowledgements.} The authors would like to thank Greig Cowan and Graham Smith of NatWest Group's Data Science \& Innovation team for the time and support needed to develop this research paper.

\bibliographystyle{apalike}
%\bibliography{ml_library}

\begingroup
    \setlength{\bibsep}{5pt}
    \setstretch{1}
    \bibliography{machineunlearning}

\begin{thebibliography}{}

\bibitem[Aldaghri et~al., 2021]{coded_2021}
Aldaghri, N., Mahdavifar, H., and Beirami, A. (2021).
\newblock Coded machine unlearning.
\newblock {\em IEEE Access}, 9:88137--88150.

\bibitem[Baumhauer et~al., 2020]{baumhauer2020machine}
Baumhauer, T., Schöttle, P., and Zeppelzauer, M. (2020).
\newblock Machine unlearning: Linear filtration for logit-based classifiers.
\newblock {\em arXiv e-prints}, page arXiv:2002.02730.

\bibitem[Biggio et~al., 2013]{biggio2013poisoning}
Biggio, B., Nelson, B., and Laskov, P. (2013).
\newblock Poisoning attacks against support vector machines.
\newblock {\em arXiv e-prints}, page arXiv:1206.6389.

\bibitem[Bourtoule et~al., 2021]{Bourtoule_2021}
Bourtoule, L., Chandrasekaran, V., Choquette-Choo, C., Jia, H., Travers, A.,
  Zhang, B., Lie, D., and Papernot, N. (2021).
\newblock Machine unlearning.
\newblock In {\em Proceedings of the 42nd IEEE Symposium on Security and
  Privacy}, San Francisco, CA.

\bibitem[Brophy and Lowd, 2021]{Brophy_2021}
Brophy, J. and Lowd, D. (2021).
\newblock Machine unlearning for random forests.
\newblock In Meila, M. and Zhang, T., editors, {\em Proceedings of the 38th
  International Conference on Machine Learning}, volume 139 of {\em Proceedings
  of Machine Learning Research}, pages 1092--1104. PMLR.

\bibitem[Cao and Yang, 2015]{CauYang}
Cao, Y. and Yang, J. (2015).
\newblock Towards making systems forget with machine unlearning.
\newblock In {\em 2015 IEEE Symposium on Security and Privacy}, pages 463--480.

\bibitem[Chen et~al., 2022]{chen2022recommender}
Chen, C., Sun, F., Zhang, M., and Ding, B. (2022).
\newblock Recommendation unlearning.
\newblock {\em arXiv e-prints}, page arXiv:2201.06820.

\bibitem[Deng et~al., 2009]{Deng_2009}
Deng, J., Dong, W., Socher, R., Li, L.-J., Li, K., and Fei-Fei, L. (2009).
\newblock Imagenet: A large-scale hierarchical image database.
\newblock {\em 2009 IEEE Conference on Computer Vision and Pattern
  Recognition}, pages 248--255.

\bibitem[Doshi-Velez and Kim, 2017]{doshivelez2017rigorous}
Doshi-Velez, F. and Kim, B. (2017).
\newblock Towards a rigorous science of interpretable machine learning.
\newblock {\em arXiv e-prints}, page arXiv:1702.08608.

\bibitem[Dwork and Roth, 2014]{Dwork_2014}
Dwork, C. and Roth, A. (2014).
\newblock The algorithmic foundations of differential privacy.
\newblock {\em Found. Trends Theor. Comput. Sci.}, 9(34):211--407.

\bibitem[Fu et~al., 2022]{fu2022knowledge}
Fu, S., He, F., and Tao, D. (2022).
\newblock Knowledge removal in sampling-based bayesian inference.
\newblock In {\em International Conference on Learning Representations}.

\bibitem[Ginart et~al., 2019]{Ginart_2019}
Ginart, A., Guan, M.~Y., Valiant, G., and Zou, J. (2019).
\newblock Making {AI} forget you: Data deletion in machine learning.
\newblock {\em arXiv e-prints}, page arXiv:1907.05012.

\bibitem[Golatkar et~al., 2020]{golatkar2020mixedprivacy}
Golatkar, A., Achille, A., Ravichandran, A., Polito, M., and Soatto, S. (2020).
\newblock Mixed-privacy forgetting in deep networks.
\newblock {\em arXiv e-prints}, page arXiv:2012.13431.

\bibitem[{Golatkar} et~al., 2019]{Golatkar_2019}
{Golatkar}, A., {Achille}, A., and {Soatto}, S. (2019).
\newblock {Eternal Sunshine of the Spotless Net: Selective Forgetting in Deep
  Networks}.
\newblock {\em arXiv e-prints}, page arXiv:1911.04933.

\bibitem[Guo et~al., 2020]{guo2020certified}
Guo, C., Goldstein, T., Hannun, A., and van~der Maaten, L. (2020).
\newblock Certified data removal from machine learning models.
\newblock {\em arXiv e-prints}, page arXiv:1911.03030.

\bibitem[Gupta et~al., 2021]{Gupta_2021}
Gupta, V., Jung, C., Neel, S., Roth, A., Sharifi-Malvajerdi, S., and Waites, C.
  (2021).
\newblock Adaptive machine unlearning.
\newblock {\em arXiv e-prints}, page arXiv:2106.04378.

\bibitem[He et~al., 2021]{he2021deepobliviate}
He, Y., Meng, G., Chen, K., He, J., and Hu, X. (2021).
\newblock {DeepObliviate: A Powerful Charm for Erasing Data Residual Memory in
  Deep Neural Networks}.
\newblock {\em arXiv e-prints}, page arXiv:2105.06209.

\bibitem[Jia et~al., 2021]{jia2021proofoflearning}
Jia, H., Yaghini, M., Choquette-Choo, C.~A., Dullerud, N., Thudi, A.,
  Chandrasekaran, V., and Papernot, N. (2021).
\newblock Proof-of-learning: Definitions and practice.
\newblock {\em arXiv e-prints}, page arXiv:2103.05633.

\bibitem[Kirkpatrick et~al., 2017]{Kirkpatrick_2017}
Kirkpatrick, J., Pascanu, R., Rabinowitz, N., Veness, J., Desjardins, G., Rusu,
  A.~A., Milan, K., Quan, J., Ramalho, T., Grabska-Barwinska, A., Hassabis, D.,
  Clopath, C., Kumaran, D., and Hadsell, R. (2017).
\newblock Overcoming catastrophic forgetting in neural networks.
\newblock {\em Proceedings of the National Academy of Sciences},
  114(13):3521--3526.

\bibitem[Koh and Liang, 2020]{koh2020understanding}
Koh, P.~W. and Liang, P. (2020).
\newblock Understanding black-box predictions via influence functions.
\newblock {\em arXiv e-prints}, page arXiv:1703.04730.

\bibitem[{Mahadevan} and {Mathioudakis}, 2021]{Mahadevan_2021}
{Mahadevan}, A. and {Mathioudakis}, M. (2021).
\newblock {Certifiable Machine Unlearning for Linear Models}.
\newblock {\em arXiv e-prints}, page arXiv:2106.15093.

\bibitem[{Neel} et~al., 2020]{Neel_2020}
{Neel}, S., {Roth}, A., and {Sharifi-Malvajerdi}, S. (2020).
\newblock {Descent-to-Delete: Gradient-Based Methods for Machine Unlearning}.
\newblock {\em arXiv e-prints}, page arXiv:2007.02923.

\bibitem[Netzer et~al., 2011]{Netzer_2011}
Netzer, Y., Wang, T., Coates, A., Bissacco, A., Wu, B., and Ng, A.~Y. (2011).
\newblock Reading digits in natural images with unsupervised feature learning.
\newblock In {\em NIPS Workshop on Deep Learning and Unsupervised Feature
  Learning 2011}.

\bibitem[{Nguyen} et~al., 2022]{Nguyen_2022}
{Nguyen}, Q.~P., {Oikawa}, R., {Mon Divakaran}, D., {Chan}, M.~C., and {Low},
  B. K.~H. (2022).
\newblock {Markov Chain Monte Carlo-Based Machine Unlearning: Unlearning What
  Needs to be Forgotten}.
\newblock {\em arXiv e-prints}, page arXiv:2202.13585.

\bibitem[Peng et~al., 1994]{Peng1994}
Peng, C.-K., Buldyrev, S.~V., Havlin, S., Simons, M., Stanley, H.~E., and
  Goldberger, A.~L. (1994).
\newblock Mosaic organization of {DNA} nucleotides.
\newblock {\em Physical Review E}, 49(2):1685--1689.

\bibitem[Quenouille, 1956]{quenouille1956}
Quenouille, M.~H. (1956).
\newblock {Notes on Bias in Estimation}.
\newblock {\em Biometrika}, 43(3-4):353--360.

\bibitem[Sakar et~al., 2018]{Sakar_2018}
Sakar, C.~O., Polat, S.~O., Katircioglu, M., and Kastro, Y. (2018).
\newblock Real-time prediction of online shoppers' purchasing intention using
  multilayer perceptron and {LSTM} recurrent neural networks.
\newblock {\em Neural Computing and Applications}, 31(10):6893--6908.

\bibitem[Schelter, 2020]{Schelter20}
Schelter, S. (2020).
\newblock Amnesia - machine learning models that can forget user data very
  fast.
\newblock In {\em CIDR 2020, 10th Conference on Innovative Data Systems
  Research, Amsterdam, The Netherlands, January 12-15, 2020, Online
  Proceedings}. www.cidrdb.org.

\bibitem[Schelter et~al., 2021]{Schelter_2021}
Schelter, S., Grafberger, S., and Dunning, T. (2021).
\newblock Hedgecut: Maintaining randomised trees for low-latency machine
  unlearning.
\newblock In {\em Proceedings of the 2021 International Conference on
  Management of Data}, SIGMOD '21, page 1545–1557, New York, NY, USA.
  Association for Computing Machinery.

\bibitem[Shafer and Vovk, 2008]{shafervovk2008conformal}
Shafer, G. and Vovk, V. (2008).
\newblock A tutorial on conformal prediction.
\newblock {\em J. Mach. Learn. Res.}, 9:371--421.

\bibitem[Sommer et~al., 2020]{sommer2020probabilistic}
Sommer, D.~M., Song, L., Wagh, S., and Mittal, P. (2020).
\newblock Towards probabilistic verification of machine unlearning.
\newblock {\em arXiv e-prints}, page arXiv:2003.04247.

\bibitem[Thudi et~al., 2021]{thudi2021necessity}
Thudi, A., Jia, H., Shumailov, I., and Papernot, N. (2021).
\newblock On the necessity of auditable algorithmic definitions for machine
  unlearning.
\newblock {\em arXiv e-prints}, page arXiv:2110.11891.

\bibitem[Vinyals et~al., 2016]{Vinyals_2016}
Vinyals, O., Blundell, C., Lillicrap, T., Kavukcuoglu, K., and Wierstra, D.
  (2016).
\newblock Matching networks for one shot learning.
\newblock In {\em Proceedings of the 30th International Conference on Neural
  Information Processing Systems}, NIPS'16, page 3637–3645, Red Hook, NY,
  USA. Curran Associates Inc.

\bibitem[Wu et~al., 2022]{Wu22PUMA}
Wu, G., Hashemi, M., and Srinivasa, C. (2022).
\newblock Puma: Performance unchanged model augmentation for training data
  removal.
\newblock In {\em Proceedings of the 36th AAAI Conference on Artificial
  Intelligence (AAAI-2022)}, Vancouver, Canada.

\bibitem[{Wu} et~al., 2020]{Wu_2020}
{Wu}, Y., {Dobriban}, E., and {Davidson}, S.~B. (2020).
\newblock {DeltaGrad: Rapid retraining of machine learning models}.
\newblock {\em arXiv e-prints}, page arXiv:2006.14755.

\bibitem[Zhang et~al., 2013]{zhang2013communication}
Zhang, Y., Duchi, J.~C., and Wainwright, M.~J. (2013).
\newblock Communication-efficient algorithms for statistical optimization.
\newblock {\em The Journal of Machine Learning Research}, 14(1):3321--3363.

\end{thebibliography}
\endgroup

\newpage
\begin{subappendices}
\renewcommand{\thesection}{\Alph{section}}%

\section{Datasets} \label{appendix:datasets}
Descriptions of the datasets included in Table \ref{table:datasets} are given below.
\begin{description}
	\item[\textbf{a.}] \textsc{mnist} is a 10-class image classification dataset containing images of digits 0--9 (\href{http://yann.lecun.com/exdb/mnist/}{http://yann.lecun.com/exdb/mnist/} [accessed: 31-August-2022]). \textsc{mnist} binary is also used in \citet{Mahadevan_2021} by taking digits 3 and 8. This has 11,982 samples with $49.00\%$ class balance.
	\item[\textbf{b.}] Covtype is a multi-class classification dataset derived from US Geological Survey and USFS involving cartographic feature variables and forest cover type as the target variable (\href{https://www.csie.ntu.edu.tw/~cjlin/libsvmtools/datasets/multiclass.html\#covtype}{https://www.csie.ntu.edu.tw/~cjlin/libsvmtools/datasets/multiclass.html\#covtype} [accessed: 31-August-2022]). \citet{Mahadevan_2021} consider the binary version of Covtype which has the same number of samples as the non-binary version with $49.00\%$ class balance (\href{https://www.csie.ntu.edu.tw/~cjlin/libsvmtools/datasets/multiclass.html\#covtype}{https://www.csie.ntu.edu.tw/~cjlin/libsvmtools/datasets/multiclass.html\#covtype} [accessed: 31-August-2022]).
	\item[\textbf{c.}] \textsc{higgs} measures kinematic properties of particle detectors in the Higgs boson accelerator and derived values as features, with whether the signal represents a Higgs boson as the target (\href{https://archive.ics.uci.edu/ml/datasets/HIGGS}{https://archive.ics.uci.edu/ml/datasets/HIGGS} [accessed: 31-August-2022]).
	\item[\textbf{d.}] Epsilon is the Epsilon dataset from the PASCAL Large Scale Learning Challenge 2008 (\href{https://www.csie.ntu.edu.tw/~cjlin/libsvmtools/datasets/binary.html\#epsilon}{https://www.csie.ntu.edu.tw/~cjlin/libsvmtools/datasets/binary.html\#epsilon} [accessed: 31-August-2022]) .
	\item[\textbf{e.}] \textsc{cifar-10/100} are 10- and 100-class image classification datasets respectively representing various animals and objects (\href{http://www.cs.toronto.edu/~kriz/cifar.html}{http://www.cs.toronto.edu/~kriz/cifar.html} [accessed: 31-August-2022]). \textsc{cifar-2} is extracted in \citet{Mahadevan_2021} from \textsc{cifar-10} by only considering the \texttt{cat} and \texttt{ship} labels. 
	\item[\textbf{f.}] \textsc{lsun} is a 10-class image classification dataset representing 10 scenes such as dining room, bedroom, and so on (\href{https://www.yf.io/p/lsun}{https://www.yf.io/p/lsun} [accessed: 31-August-2022]) .
	\item[\textbf{g.}] \textsc{sst} is an NLP dataset consisting of movie reviews for sentiment analysis (\href{https://nlp.stanford.edu/sentiment/index.html}{https://nlp.stanford.edu/sentiment/index.html} [accessed: 31-August-2022]) .
	\item[\textbf{h.}] \textsc{svhn} is a 10-class image classification dataset consisting of house numbers taken from Google Street View, where the objective is to identify the digits 0--9 (\href{http://ufldl.stanford.edu/housenumbers/}{http://ufldl.stanford.edu/housenumbers/} [accessed: 31-August-2022]).
	\item[\textbf{i.}] \textsc{rcv1} is a collection of manually labelled news articles from Reuters taken from the period 1996-1997 (\href{https://www.csie.ntu.edu.tw/~cjlin/libsvmtools/datasets/binary.html\#rcv1.binary}{https://www.csie.ntu.edu.tw/~cjlin/libsvmtools/datasets/binary.html\#rcv1.binary} [accessed: 31-August-2022]). 
	\item[\textbf{j.}] Purchase consists of consumer purchase history with the target variable being whether a customer will repeat purchase (\href{https://www.kaggle.com/c/acquire-valued-shoppers-challenge/data}{https://www.kaggle.com/c/acquire-valued-shoppers-challenge/data} [accessed: 31-August-2022]). In \citet{Bourtoule_2021}, Purchase is curated in \citet{Bourtoule_2021} from the original Purchase dataset by choosing the top 600 most purchased items based on the \texttt{category} attribute.
	\item[\textbf{k.}] ImageNet is a large-scale image dataset which have been annotated according to the WordNet hierarchy, with the 1000-class target variable being the annotation's synset (\href{https://www.image-net.org/}{https://www.image-net.org/} [accessed: 31-August-2022]). Mini-ImageNet is a supervised classification version of ImageNet created by the process of \citet{Vinyals_2016}.
	\item[\textbf{l.}] Surgical is a binary-classification dataset with the goal of predicting whether a surgery involved complications (\href{https://www.kaggle.com/datasets/omnamahshivai/surgical-dataset-binary-classification}{https://www.kaggle.com/datasets/omnamahshivai/surgical-dataset-binary-classification} [accessed: 31-August-2022]).
	\item[\textbf{m.}] Vaccine is a binary-classification dataset on whether a person had got a flu vaccine (\href{https://www.drivendata.org/competitions/66/flu-shot-learning/}{https://www.drivendata.org/competitions/66/flu-shot-learning/} [accessed: 31-August-2022]).
	\item[\textbf{n.}] Adult is a binary-classification dataset determining whether a person has an annual income of over \$50,000 (\href{http://archive.ics.uci.edu/ml/datasets/Adult}{http://archive.ics.uci.edu/ml/datasets/Adult} [accessed: 31-August-2022]).
	\item[\textbf{o.}] Bank Mktg. consists of a Portuguese bank marketing call details with whether a contacted customer subsequently subscribed as the target variable (\href{http://archive.ics.uci.edu/ml/datasets/Bank+Marketing}{http://archive.ics.uci.edu/ml/datasets/Bank+Marketing} [accessed: 31-August-2022]).
	\item[\textbf{p.}] Diabetes is a binary-classification dataset of diabetic patients whose target variable is hospital readmission (\href{https://archive.ics.uci.edu/ml/datasets/diabetes}{https://archive.ics.uci.edu/ml/datasets/diabetes} [accessed: 31-August-2022]).
	\item[\textbf{q.}] No Show is a binary-classification dataset with whether a patient missed a doctor appointment as the target variable (\href{https://www.kaggle.com/datasets/joniarroba/noshowappointments}{https://www.kaggle.com/datasets/joniarroba/noshowappointments} [accessed: 31-August-2022]).
	\item[\textbf{r.}] Olympic is a binary-classification dataset for predicting whether an athlete received an Olympic medal for the event participated (\href{https://www.kaggle.com/datasets/heesoo37/120-years-of-olympic-history-athletes-and-results}{https://www.kaggle.com/datasets/heesoo37/120-years-of-olympic-history-athletes-and-results} [accessed: 31-August-2022]).
	\item[\textbf{s.}] Census is a binary-classification dataset for predicting whether a person has an annual income of over \$50,000 based on census data (\href{https://archive.ics.uci.edu/ml/datasets/Census-Income+(KDD)}{https://archive.ics.uci.edu/ml/datasets/Census-Income+(KDD)} [accessed: 31-August-2022]).
	\item[\textbf{t.}] Credit Card is a highly-imbalanced binary-classification dataset for predicting fraudulent European credit card transactions in September 2013 (\href{https://www.kaggle.com/datasets/mlg-ulb/creditcardfraud}{https://www.kaggle.com/datasets/mlg-ulb/creditcardfraud} [accessed: 31-August-2022]).
	\item[\textbf{u.}] \textsc{ctr} is curated by \cite{Brophy_2021} by taking the first 1,000,000 instances and 13 numeric attributes from Criteo's original dataset; each row represents an ad that was displayed and the binary target variable indicates whether the ad was clicked on (\href{https://ailab.criteo.com/download-criteo-1tb-click-logs-dataset/}{https://ailab.criteo.com/download-criteo-1tb-click-logs-dataset/} [accessed: 31-August-2022]).
	\item[\textbf{v.}] Twitter is curated by \cite{Brophy_2021} by taking the first 1,000,000 tweets from the HSpam14 dataset (\href{https://www3.ntu.edu.sg/home/AXSun/datasets.html}{https://www3.ntu.edu.sg/home/AXSun/datasets.html} [accessed: 31-August-2022]). This dataset has a binary target variable indicating whether a tweet is spam or not.
	\item[\textbf{w.}] Lacuna-10/100 are 10- and 100-class image classification datasets consisting of the faces of 10 and 100 different celebrities, respectively, extracted by \cite{Golatkar_2019} from the VGGFaces2 dataset (\href{https://www.robots.ox.ac.uk/ vgg/data/vgg\_face2/}{https://www.robots.ox.ac.uk/vgg/data/vgg\_face2/} [accessed: 31-August-2022]).
\end{description}

\end{subappendices}

\end{document}